%% file: main.tex
\documentclass[11pt,letterpaper,logo,onecolumn,copyright]{hunyuan_report}

\usepackage[authoryear,sort&compress,round]{natbib}
\input{math_commands.tex}

\usepackage{array,multirow}
\usepackage{microtype}
\usepackage[breakable,skins]{tcolorbox}
\usepackage{fvextra}
\usepackage{flafter}
\usepackage{placeins}

\hypersetup{breaklinks=true,hypertexnames=false}
\colorlet{takeawayborder}{gblue9!15}
\colorlet{takeawaytitle}{gblue9!8}
\colorlet{takeawaybody}{gblue9!5}
\newcommand{\ours}{truncLongPPL}
\newcommand{\best}[1]{\textbf{#1}}

\newcolumntype{L}[1]{>{\raggedright\arraybackslash}p{#1}}
\newcolumntype{C}[1]{>{\centering\arraybackslash}p{#1}}
\newtcolorbox{takeawaybox}[1]{
  enhanced,
  breakable,
  colback=gblue9!5,
  colframe=gblue9!15,
  colbacktitle=gblue9!8,
  coltitle=gblue9,
  fonttitle=\bfseries,
  title={#1},
  boxrule=0.4pt,
  arc=6pt,
  left=12pt,
  right=12pt,
  top=8pt,
  bottom=8pt,
  toptitle=5pt,
  bottomtitle=5pt,
  before skip=0.8em,
  after skip=1.0em
}

\title{Which Long Documents Matter?\\A Controlled Study of Data Selection for Context Extension}
\runningtitle{Which Long Documents Matter? A Controlled Study of Data Selection for Context Extension}
\author{
Junlong Jia, Jiang Zhou, Xing Wu\textsuperscript{\textdagger}, Chaochen Gao\\
\bfseries Tinghao Yu\textsuperscript{\textdagger}, Feng Zhang, Lilin Wang\\
\vspace{0.3cm}
\normalfont\normalsize Hunyuan Team, Tencent\\
{\normalfont\fontfamily{lmtt}\selectfont \{ucaswu,maxwellyu\}@tencent.com}
}

\begin{abstract}
Natural long documents are widely used for context extension, but which selection pipelines are effective remains unclear. We conduct a controlled comparison on Qwen2.5-3B at 128K under matched 6B-token continued-training budgets. Random sampling of natural long documents (random-long) outperforms repacking at longer evaluation lengths. Within the same candidate pool, selecting documents with high LongPPL key-token density yields further gains, unlike the evaluated generic-quality and LLM-judge selectors. We introduce truncLongPPL to reduce scoring cost by computing key-token density over a truncated prefix. At 128K, scoring the first half of the reference span yields downstream performance close to full-span scoring with a measured $2.28\times$ end-to-end scoring speedup. Across three training seeds, truncLongPPL achieves the highest mean scores among evaluated pipelines, exceeding random-long by 4.54 points on validation HELMET and 3.28 on held-out RULER, with gains most consistent in retrieval. At a 256K target context, it gains 2.82 average RULER points over random-long on Qwen2.5-3B. In short/long mixed training across Qwen2.5-3B, Qwen3-8B, and OLMo~3-7B, it gains 1.47--7.31 average RULER points over random-long mixing, with seven-task short-context averages within 0.23 points of the corresponding baselines. After supervised fine-tuning (SFT), Qwen3-8B and OLMo~3-7B models trained with selected long documents outperform their random-long counterparts by 1.13--1.67 points on LongBench v2 and 0.62--1.44 points on MRCR.
\end{abstract}

\begin{document}
\maketitle
\begingroup
\renewcommand{\thefootnote}{\textdagger}
\footnotetext[2]{Corresponding authors.}
\endgroup
\setcounter{footnote}{0}

\section{Introduction}
\label{sec:intro}

Long-context extension commonly involves continued training at longer sequence lengths \citep{gao2024prolong,yang2025qwen3,liu2024deepseek,sun2026data}. Natural long documents are useful for this stage because they can contain cross-section structure, delayed references, and distributed evidence \citep{gao2024prolong,olmo2025olmo}. However, length alone does not guarantee useful supervision: some documents contain substantial cross-passage dependencies, whereas others largely consist of locally coherent but independent passages \citep{deng2025longfilter,chen2024prospector,fang2025wrong}. This raises two practical questions: do natural long documents offer benefits over repacking, and which selection pipelines improve on random sampling?

Existing selection pipelines use different signals, ranging from generic-quality proxies such as perplexity (PPL) and gzip compressibility to dependency-aware signals based on model judgments or likelihood changes induced by additional context \citep{olmo2025olmo,deng2025longfilter,chen2024prospector,fang2025wrong}. Their findings are difficult to compare because corpora, scoring models, target contexts, training budgets, and mixture ratios vary across studies. OLMo~3, for example, evaluates document-level filters derived from LongPPL key-token statistics but reports no clear advantage over gzip in its setting \citep{olmo2025olmo}. These differences in experimental settings motivate a controlled comparison under shared training and evaluation conditions.

We conduct this comparison on Qwen2.5-3B \citep{yang2024qwen25} at 128K with matched 6B-token continued-training budgets. We first compare repacking with random sampling of natural long documents, then evaluate generic-quality, LLM-judge, and likelihood-based selection pipelines within a fixed natural-long candidate pool. HELMET \citep{yen2024helmet} determines the selection configuration, while RULER \citep{hsieh2024ruler} is held out for evaluation; seven short-context tasks assess capability retention. These comparisons evaluate complete pipelines and identify LongPPL-based key-token density as a promising selection criterion, motivating a closer examination of its selectivity and computational cost.

To understand when LongPPL-based selection helps, we first examine how the retained density range and filtering strength affect downstream performance. Full-span scoring is expensive at corpus scale, so we then test whether shorter prefixes preserve the selection benefit at lower cost. This leads to \ours{}, which computes key-token density over a truncated prefix. In the controlled 128K study, front-half scoring yields downstream scores close to full-span scoring while reducing scoring cost. Capability-level analysis further shows that the most consistent gains are in retrieval, with less uniform benefits for other long-context capabilities.

We further validate selection at a 256K target context on Qwen2.5-3B under the same training-token budget. The short-context decline observed during continued training also motivates experiments that mix high-quality short data with natural long documents. We compare random and selected long data within these mixtures, extend the evaluation to Qwen3-8B \citep{yang2025qwen3} and OLMo~3-7B \citep{olmo2025olmo}, and use LongBench v2 and MRCR to assess whether the selection gains persist after supervised fine-tuning (SFT).

\begin{takeawaybox}{Key findings}
\small
\begin{itemize}[leftmargin=1.5em,itemsep=0.35em,topsep=0.35em]
    
    \item \textbf{Natural long documents outperform repacking
    at longer evaluation lengths.}
    In our controlled comparison, random sampling of natural
    long documents yields higher scores than repacking,
    with smaller differences at shorter evaluation lengths
    and larger gains at longer lengths
    (\S\ref{sec:natural-vs-repack}).

    \item \textbf{Likelihood-based dependency selection yields
    further gains.}
    Within the same natural-long candidate pool, the evaluated
    likelihood-based long-dependency pipelines outperform
    random-long, whereas generic-quality and LLM-judge
    selection do not. \ours{} achieves the highest mean
    HELMET and RULER scores among the evaluated pipelines
    across three training seeds
    (\S\ref{sec:signals}).

    \item \textbf{Front-half scoring reduces cost with similar
    downstream performance.}
    In the controlled study, \ours{} scores only the first
    half of the reference span and yields downstream scores
    close to full-span scoring, with a measured $2.28\times$
    end-to-end scoring speedup
    (\S\ref{sec:robustness}).

    \item \textbf{Selection gains are most consistent in retrieval.}
    Compared with random-long, \ours{} improves retrieval
    on both HELMET and RULER, with larger gains at longer
    evaluation lengths. QA effects depend on the benchmark,
    while RULER aggregation and variable tracking show
    no consistent improvement
    (\S\ref{sec:scope}).

    \item \textbf{Selection gains extend to longer target contexts.}
    Extending the controlled comparison from a 128K to a 256K
    target context preserves the advantage of \ours{} over
    random-long
    (\S\ref{sec:length-transfer}).

    \item \textbf{Selection also benefits mixed training
    and subsequent SFT.}
    When selected long documents are combined with short data,
    long-context scores improve across the tested models,
    while seven-task short-context averages remain close
    to those of random-long mixing. The overall selection
    advantage also persists in the post-SFT evaluations
    (\S\ref{sec:olmo-recipe}).




\end{itemize}
\end{takeawaybox}

\section{Experimental Setup and Evaluation Protocol}
\label{sec:setup}

\subsection{Study Design and Training Settings}

The pipeline comparisons and sensitivity and capability analyses use Qwen2.5-3B at 128K with 6B continued-training tokens per configuration. Length-extension experiments keep the same backbone and token budget and increase the target context to 256K. Mixed-training experiments use Qwen2.5-3B at 128K with 10B tokens and a 66/34 short/long mixture, and Qwen3-8B Base and OLMo~3-7B at 256K with 20B tokens and a 66/34 mixture. Table~\ref{tab:protocol} summarizes these settings. Within each setting, the compared pipelines use matched training budgets.

\begin{table}[!htbp]
\caption{Overview of experiments and continued-training settings. Mixture ratios are short:long. The dolma3-short controls pack only short documents and use the same total budget as their respective mixed configurations. The short-data source is described in Appendix~\ref{app:short-data}.}
\label{tab:protocol}
\centering
\small
\setlength{\tabcolsep}{4pt}
\begin{tabular}{@{}lllll@{}}
\toprule
Study & Backbone & Context & Tokens & Data setting \\
\midrule
Controlled comparison & Qwen2.5-3B & 128K & 6B & Controlled pipelines \\
Length extension & Qwen2.5-3B & 128K / 256K & 6B & Natural long \\
Mixture validation & Qwen2.5-3B & 128K & 10B & Mixed, 66/34 \\
Cross-model validation & Qwen3-8B & 256K & 20B & Mixed, 66/34 \\
Cross-model validation & OLMo~3-7B & 256K & 20B & Mixed, 66/34 \\
\bottomrule
\end{tabular}
\end{table}

\paragraph{Mixed-training validation.}
The mixed-training studies retain high-quality short data to balance long-context extension with short-context capability retention. Within each setting, we compare random-long with the default \ours{} configuration (Section~\ref{sec:prefix-selection}) while keeping the total token budget and short/long ratio fixed. The dolma3-short configuration provides a reference for the contribution of long data.

\paragraph{Optimization for the controlled Qwen2.5-3B 128K study.}
We extend Qwen2.5-3B directly to 128K by increasing the RoPE base to $\theta=2\times10^{8}$. Training uses bf16 and AdamW ($\beta_1{=}0.9$, $\beta_2{=}0.95$, weight decay $0.1$, gradient clip $1.0$), a constant learning rate of $1\times10^{-6}$ without warmup or decay, and approximately 8M tokens per optimizer update.

\subsection{Task, Data, and Selection Pipelines}
\label{sec:data-methods}

\paragraph{Task and source data.}
A selector assigns a score $s(x)$ to each natural long document in a candidate pool $\mathcal{C}$ and produces a subset $\mathcal{D}\subseteq\mathcal{C}$ for continued training. We compare the resulting training pipelines by downstream performance under matched token budgets within each setting. The research corpus is sampled from 25\% of the documents in our internal pretraining corpus, covering books, scanned documents, manuals, academic papers, and web sources in English, Chinese, and mixed/other languages. The Qwen2.5-3B 128K study retains 3M documents of at least 128K tokens. The corpus excludes repository-level code.

\paragraph{Controlled pipeline comparison.}
The Qwen2.5-3B 128K study compares repacking, random-long, PPL- and gzip-based generic-quality selectors, the LLM-judge selector, LongFilter, Prospector, and LongPPL-density selection. All natural-long selectors in that study operate on the same candidate pool. Table~\ref{tab:data-methods} and Appendix~\ref{app:pipeline-configurations} give their constructions, scoring models, and selection rules. The 256K length-extension comparison evaluates \ours{} against random-long and LongFilter.

\subsection{LongPPL-Density Selection and Prefix Scoring}
\label{sec:prefix-selection}

We select documents with a high proportion of tokens that become substantially more predictable with long context. Prefix scoring limits the text processed to compute this document-level score.

Following LongPPL \citep{fang2025wrong}, let $k_i\in\{0,1\}$ indicate whether token $x_i$ is key: its long-context log-likelihood must exceed the short-context reference by more than $\alpha$ and exceed $\beta$. Let $L$ be the reference span, set to the nominal target-context length, and let $r$ specify the prefix fraction. To reduce the cost of full-span scoring, we truncate the scoring input at $T_r=\lfloor rL\rfloor$ and call this prefix-based variant \ours{}. The first $K=4096$ tokens provide initial context and are excluded from the density calculation. The valid target positions and key-token density are
\begin{equation}
    W_r=\{K+1,\ldots,T_r\},\qquad
    \rho_{W_r}(x)=\frac{1}{T_r-K}\sum_{i=K+1}^{T_r}k_i.
    \label{eq:prefix-density}
\end{equation}
We retain a document when $\rho_{W_r}(x)>\tau$. The default is $r=1/2$. Fractions $r=1/2$ and $r=1/4$ give prefix endpoints of 64K and 32K when $L=128$K, and 128K and 64K when $L=256$K. The fraction is taken relative to $L$, not to each raw document's variable length, and positions are counted with the scoring model's tokenizer. Appendix~\ref{app:likelihood-details} gives the token-level definitions and scoring details; Section~\ref{sec:robustness} reports the controlled Qwen2.5-3B scoring study.

\subsection{Evaluation and Uncertainty}

\paragraph{Long-context evaluation.}
For Qwen2.5-3B, the 128K setting evaluates RULER \citep{hsieh2024ruler} at 8K, 16K, 32K, 64K, and 128K; the 256K length-extension setting additionally includes 256K. For Qwen3-8B and OLMo~3-7B mixed training, we report RULER over 8K--256K and at the 256K target length. Full-range averages cover different length sets, so selection gains are compared against baselines within each setting.

In the Qwen2.5-3B 128K selection analyses, HELMET \citep{yen2024helmet} determines the density threshold and scoring-window configuration; RULER is evaluated after those choices are fixed. HELMET is the unweighted mean of recall, in-context learning, re-ranking, long-document question answering (QA), and retrieval-augmented generation (RAG). The capability analysis decomposes RULER into retrieval, aggregation, variable tracking, and QA.

\paragraph{Short-context evaluation.}
We use \texttt{lm-evaluation-harness} \citep{gao2024lmeval} in a zero-shot setting on HellaSwag \citep{zellers2019hellaswag}, LAMBADA (OpenAI) \citep{paperno2016lambada}, ARC-Challenge and ARC-Easy \citep{clark2018arc}, PIQA \citep{bisk2020piqa}, WinoGrande \citep{sakaguchi2020winogrande}, and LogiQA \citep{liu2020logiqa}. The seven-task short-context average is the unweighted mean of the seven raw accuracy metrics.

\paragraph{Repeats and uncertainty.}
The nine-pipeline Qwen2.5-3B 128K comparison reports means and standard deviations across three training seeds (Figure~\ref{fig:signals}, Appendix~\ref{app:main-seeds}). The length-specific, capability, threshold, window, and budget analyses use one reference training run per configuration. The Qwen2.5-3B fixed-mixture comparison reports three independent training runs.

\section{Controlled Pipeline Comparison on Qwen2.5-3B}
\label{sec:selection}

We study data construction and selection on Qwen2.5-3B at 128K under a shared 6B-token budget. We first compare repacking with random sampling of natural long documents. We then evaluate whether document-selection pipelines improve on random sampling within the same natural-long candidate pool.

\subsection{Natural Long-Document Sampling versus Repacking}
\label{sec:natural-vs-repack}

The \textsc{repack} baseline packs documents into 128K training sequences, following the repacking strategy of MAI-Thinking-1 \citep{mai2026thinking}. We compare it with random-long (labeled random in the figures), which uniformly samples natural long documents without quality scoring. Both pipelines follow the constructions in Appendix~\ref{app:pipeline-configurations} and use the same training-token budget.

\begin{figure}[!tb]
\centering
\includegraphics[width=0.80\linewidth]{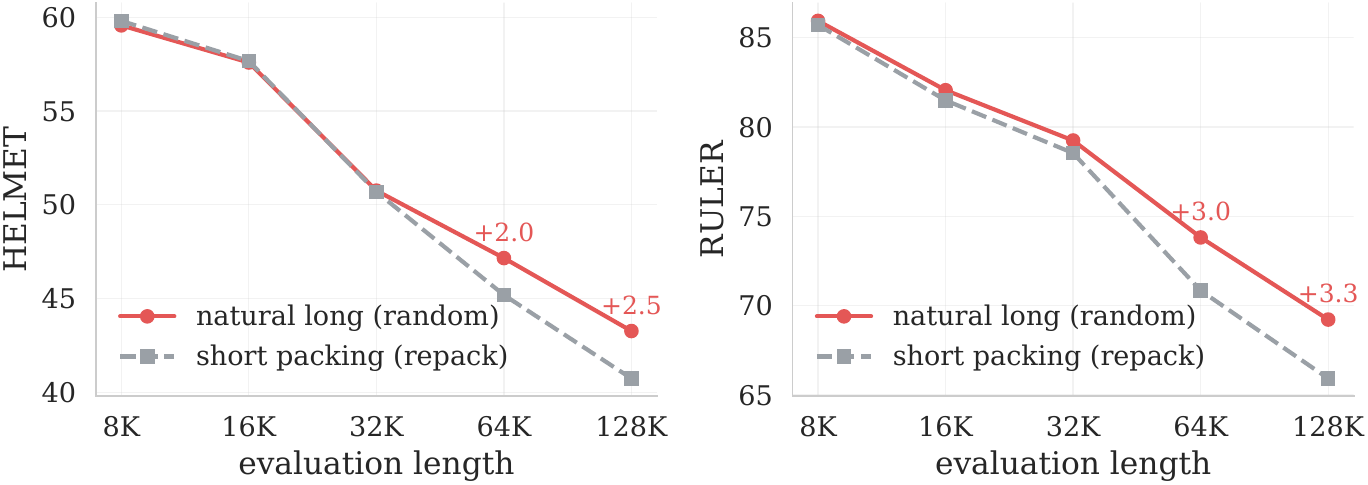}
\caption{Validation HELMET (left) and held-out RULER (right) by evaluation length for random sampling of natural long documents and \textsc{repack} under a shared 6B-token budget. At 128K, random sampling is higher by approximately 2.5 and 3.3 points, respectively.}
\label{fig:natural-vs-repack}
\end{figure}

Figure~\ref{fig:natural-vs-repack} shows only a small difference at 8K, whereas the gap increases with evaluation length. At 128K, random sampling of natural long documents is higher than \textsc{repack} by approximately 2.5 validation HELMET points and 3.3 held-out RULER points. This growing gap motivates examining how to select training data within the natural-long candidate pool.\par

\subsection{Pipelines for Selecting Natural Long Documents}
\label{sec:signals}

\begin{figure}[!tb]
\centering
\includegraphics[width=\linewidth]{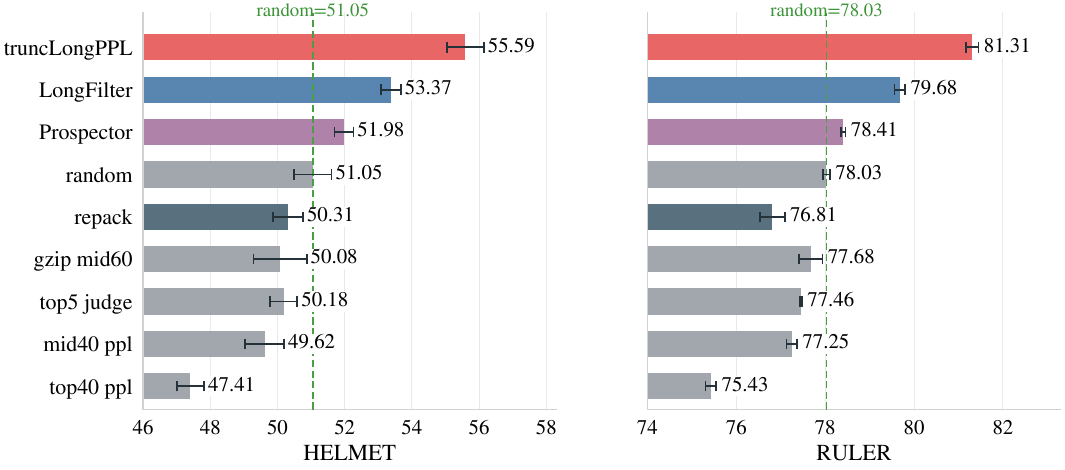}
\caption{Three-seed comparison of data-selection pipelines at 128K and 6B training tokens. Bars show mean validation HELMET (left) and held-out RULER (right); error bars show one standard deviation across training seeds. Dashed lines mark the random-long means. \ours{} uses $r=1/2$ prefix scoring at $\tau=0.05$; exact values appear in Appendix~\ref{app:main-seeds}.}
\label{fig:signals}
\end{figure}

We next compare the complete pipelines defined in Appendix~\ref{app:pipeline-configurations}: repacking and random sampling as scoring-free references; PPL- and gzip-based generic-quality selectors; the LLM-judge selector; and the likelihood-based LongFilter, Prospector, and LongPPL-density selectors. Figure~\ref{fig:signals} reports \ours{}, the $r=1/2$ prefix-scoring configuration with $\tau=0.05$, which is selected using HELMET validation and subsequently evaluated on held-out RULER. Section~\ref{sec:robustness} analyzes the threshold and scoring span.\par

None of the evaluated PPL-, gzip-, or LLM-judge-based pipelines scores higher than random sampling of natural long documents; the top-40\% PPL pipeline yields the lowest scores on both long-context aggregates. To examine how the selection convention affects the comparison, we also evaluate LongPPL-density selection and gzip selection using an OLMo-style symmetric middle-60\% rule. Their scores differ only slightly under this control (Appendix~\ref{app:olmo-reconcile}), consistent with the OLMo~3 finding.\par

Both prior long-dependency pipelines score higher than random sampling. Across three training seeds, LongFilter reaches 53.37\,$\pm$\,0.30 HELMET / 79.68\,$\pm$\,0.12 RULER, and our Prospector adaptation reaches 51.98\,$\pm$\,0.28 / 78.41\,$\pm$\,0.05, compared with 51.05\,$\pm$\,0.56 / 78.03\,$\pm$\,0.08 for random sampling.

The HELMET-selected \ours{} pipeline achieves the highest mean scores at 55.59\,$\pm$\,0.55 / 81.31\,$\pm$\,0.14, improving on random sampling by 4.54 / 3.28 points and on LongFilter by 2.22 / 1.63 points. Appendix~\ref{app:main-seeds} reports the repeated runs. These results motivate examining how selective the density criterion should be and what fraction of the reference span must be scored to retain the downstream benefit.

\section{LongPPL-Density Selection: Sensitivity and Efficiency}
\label{sec:robustness}

We analyze the selection threshold and scoring window to balance downstream performance against scoring cost. HELMET validation first determines the threshold with full-span scoring ($r=1$, $L=128$K); we then shorten the window and compare its placement.

\subsection{Sensitivity to the Key-Token-Density Threshold}
\label{sec:threshold-robustness}

\begin{figure}[!htbp]
\centering
\includegraphics[width=0.80\linewidth]{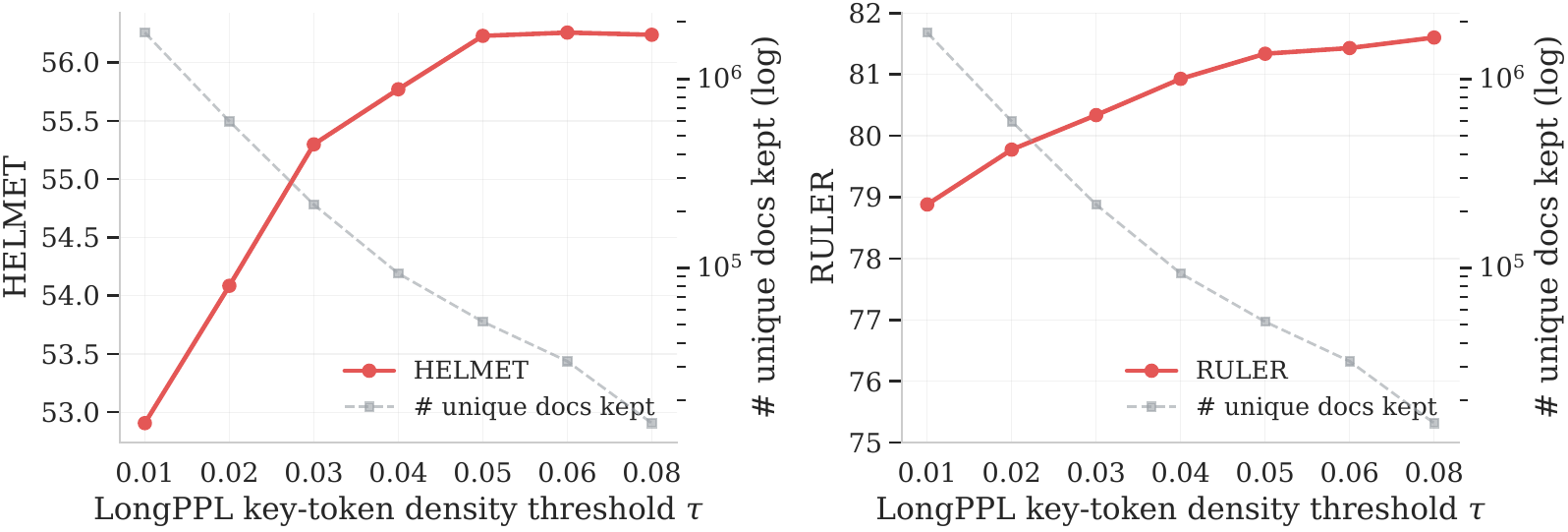}
\caption{Full-span ($r=1$, $L=128$K) LongPPL key-token-density threshold sweep under the 6B-token budget: macro-averaged validation HELMET (left), held-out RULER (right), and the number of unique documents retained (gray, secondary axes). Validation performance rises through $\tau=0.05$ and then remains on a plateau through $0.08$ as the retained pool continues to shrink.}
\label{fig:threshold-robustness}
\end{figure}

With full-span scoring ($r=1$, $L=128$K), HELMET rises substantially from $\tau=0.01$ to $0.03$, increases from 55.30 to 56.23 between $\tau=0.03$ and $0.05$, and then plateaus through $\tau=0.08$. We choose $\tau=0.05$, which retains the largest document pool on this plateau. RULER follows the same trend: approximately 78.9 at $\tau=0.01$, 80.34 at $0.03$, 81.34 at $0.05$, and approximately 81.6 at $0.08$. Removing low-density documents improves performance, with little further gain at stricter thresholds.

\subsection{Sensitivity to Scoring-Window Length}
\label{sec:window-length}

Full-span scoring requires on the order of $10^4$--$10^5$ H20 GPU-hours for the candidate pool (Appendix~\ref{app:likelihood-details}). We fix $\tau=0.05$ and compare prefix fractions $r\in\{1/8,1/4,1/2,1\}$. At $L=128$K, these fractions correspond to scoring-input endpoints of 16K, 32K, 64K, and 128K. Shortening the prefix reduces the sequence length processed by the scoring model and therefore the scoring cost.

\begin{figure}[!tb]
\centering
\includegraphics[width=0.80\linewidth]{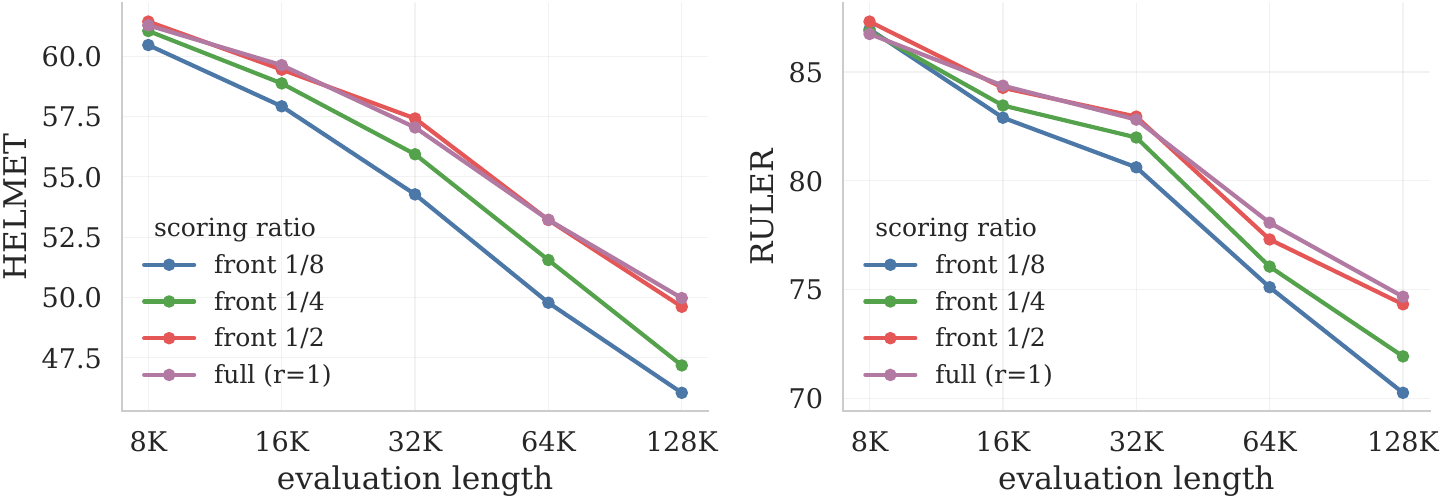}
\caption{Scores by evaluation length for front-prefix fractions $r\in\{1/8,1/4,1/2,1\}$ of $L=128$K at $\tau=0.05$ under the 6B-token budget: validation HELMET (left) and held-out RULER (right). At the 128K evaluation length, $r=1/2$ and $r=1$ are comparable, differing by 0.1 point on HELMET and 0.2 point on RULER.}
\label{fig:window-eval-length}
\end{figure}

Window length matters most at longer evaluation lengths (Figure~\ref{fig:window-eval-length}). At 128K, HELMET scores are 46.1, 47.2, 49.9, and 50.0 for $r=1/8$, $1/4$, $1/2$, and $1$, respectively. In this reference-run ablation, $r=1/2$ and $r=1$ also share a HELMET macro-average of 56.23. We choose $r=1/2$ as the shortest tested fraction with performance comparable to full-span scoring. On held-out RULER, the two differ by only 0.2 point at 128K, while $r=1/8$ reaches 70.3.

Front-half and tail-half windows have similar HELMET scores, with no advantage for the tail (Appendix~\ref{app:window-position}). The front-half scoring window is cheaper because it limits the scoring input to $L/2$, whereas tail-window scoring still requires the preceding context. Across $\tau\in\{0.03,0.04,0.05\}$, eight scattered windows with a total nominal span of $L/2$ differ from the contiguous front-half scoring window by approximately one HELMET point or less, with no consistent winner (Appendix~\ref{app:scattered}). The intersection of the document sets selected by the front-half and tail-half windows also does not outperform front-half selection at the default threshold. We use the front-half scoring prefix for \ours{}.

\subsection{Scoring Efficiency and Agreement}
At 128K, front-half scoring reduces per-document time from $16.55$ to $7.25$ seconds, a measured $2.28\times$ end-to-end speedup (Appendix~\ref{app:agreement-cost}). The speedup exceeds $2\times$ because halving the processed prefix length also reduces the superlinear attention cost in the full-context stage.

In the audited subset, key-token densities from front-half and full-span scoring are strongly correlated, and $76\%$--$92\%$ of documents selected by front-half scoring are also selected by full-span scoring. Their selected sets overlap substantially while downstream performance remains similar. We adopt the front-half configuration at $\tau=0.05$; additional budget and window analyses appear in Appendix~\ref{app:window-robustness}.

\section{Capability-Level Effects of LongPPL-Density Selection}
\label{sec:scope}

The aggregate comparison shows that LongPPL-density selection improves overall long-context performance. To identify which capabilities account for these gains and how they vary with evaluation length, we compare \ours{} with random-long by capability and evaluation length on HELMET and RULER.

\begin{figure}[!tb]
\centering
\includegraphics[width=\linewidth]{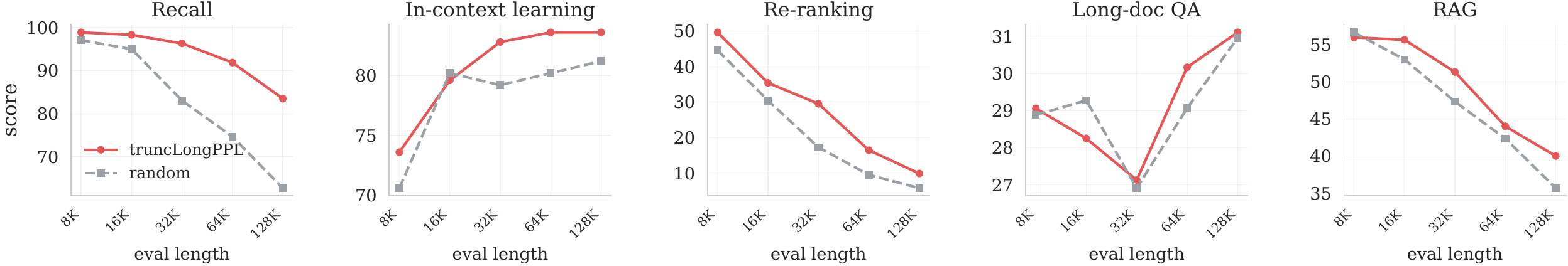}
\caption{Validation HELMET category scores by evaluation length for random sampling of natural long documents and \ours{} under the 6B-token budget. At 128K, recall is approximately 21 points higher with selection; long-document QA differs by at most 1.1 points across lengths.}
\label{fig:helmet-categories}
\end{figure}

\begin{figure}[!tb]
\centering
\includegraphics[width=\linewidth]{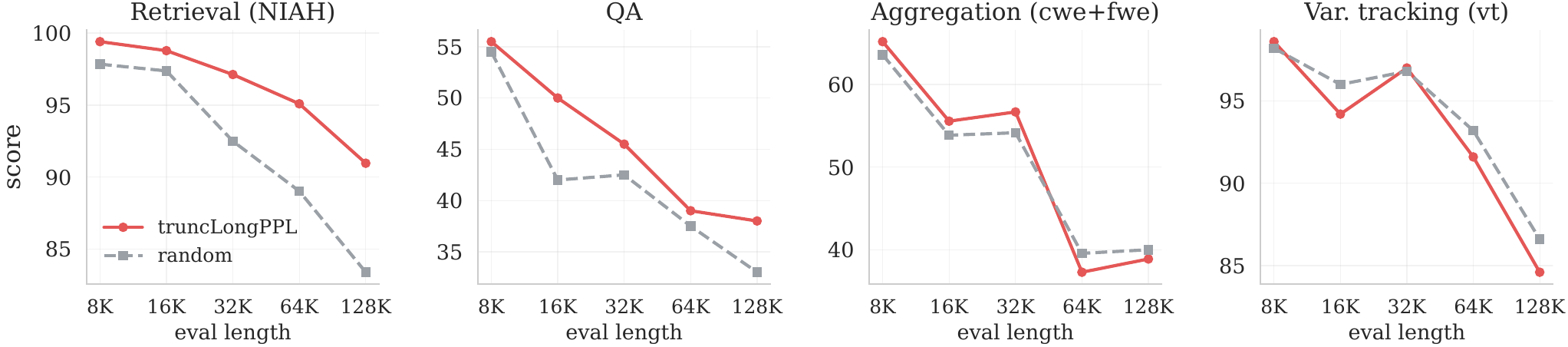}
\caption{Held-out RULER category scores by evaluation length for random sampling of natural long documents and \ours{} under the 6B-token budget.}
\label{fig:ruler-categories}
\end{figure}

\paragraph{Validation HELMET categories.}
HELMET recall shows the largest gain (Figure~\ref{fig:helmet-categories}). The gap grows from 1.8 points at 8K ($98.9$ vs.\ $97.1$) to approximately $21$ points at 128K ($83.5$ vs.\ $62.8$). Re-ranking, RAG, and in-context learning improve by approximately $4$, $4$, and $2$ points at 128K, respectively. Long-document QA differs by at most 1.1 points across lengths.

\paragraph{Held-out RULER categories.}
RULER retrieval also improves most consistently, with a small gap at 8K widening to approximately eight points at 128K (Figure~\ref{fig:ruler-categories}). QA improves at every length, but the gap varies: approximately eight points at 16K, five at 128K, and smaller differences at 32K and 64K. Aggregation and variable tracking show no consistent gain, and the \ours{} model is slightly below random-long at 128K on these categories.

The shared pattern is a retrieval gain; QA depends on the benchmark, while aggregation and variable tracking change little. We next test whether the selection benefit persists at a 256K target context.

\FloatBarrier
\section{Length Extension to 256K on Qwen2.5-3B}
\label{sec:length-transfer}

We evaluate natural-long-document selection on Qwen2.5-3B at a 256K target context, using the same 6B-token budget per configuration as in the 128K study. At each target length, we compare \ours{} with random-long and LongFilter. Additional scoring-window comparisons appear in Appendix~\ref{app:length-window}.

\begin{table}[!htbp]
\caption{Length-extension validation on Qwen2.5-3B with 6B training tokens per configuration. The 128K column reports RULER mean $\pm$ standard deviation over three training seeds. The 128K average covers 8K--128K; the 256K columns report the average over 8K--256K and the 256K endpoint. These averages cover different length sets.}
\label{tab:qwen25-256k}
\centering
\small
\setlength{\tabcolsep}{5pt}
\begin{tabular}{@{}lcrr@{}}
\toprule
& 128K training & \multicolumn{2}{c}{256K training} \\
\cmidrule(lr){2-2}\cmidrule(lr){3-4}
Selection pipeline & RULER & RULER & RULER 256K \\
\midrule
random-long & $78.03\pm0.08$ & 75.81 & 62.91 \\
LongFilter & $79.68\pm0.12$ & 77.25 & 65.75 \\
\ours{} & $\mathbf{81.31\pm0.14}$ & \textbf{78.63} & \textbf{67.05} \\
\bottomrule
\end{tabular}
\end{table}

At 128K, \ours{} exceeds random-long and LongFilter by 3.28 and 1.63 RULER points. At 256K, the gains are 2.82 and 1.38 points on the average, and 4.14 and 1.30 points at the 256K endpoint. Selection remains beneficial at the longer target context.

\FloatBarrier
\section{Selection in Short/Long Mixed Training}
\label{sec:olmo-recipe}

Long-context training also needs to preserve short-context capabilities. In the 128K study, the seven-task short-context average falls from 60.1 for the base model to 59.0--59.5 after continued training. The \ours{} model scores 59.2, down 0.9 point (Appendix~\ref{app:short-performance}). We therefore compare random and selected natural long documents mixed with high-quality short data at matched budgets and mixture ratios.

\subsection{Fixed-Mixture and Ratio Validation on Qwen2.5-3B}
\label{sec:mixture-qwen25}

We first study the mixture on Qwen2.5-3B, then test Qwen3-8B and OLMo~3-7B at 256K and evaluate the models after SFT.
The Qwen2.5-3B comparison uses a 66/34 short/long ratio, following OLMo~3, and a separate \textbf{10B}-token budget: 6.6B short tokens and 3.4B long tokens in the mixtures. The random-long and \ours{} mixtures share the short-data component and natural-long candidate pool. We also test 50/50 and 34/66 ratios (Appendix~\ref{app:mixture-ratios}). The three 128K configurations are:
\begin{itemize}[leftmargin=1.5em,itemsep=0.35em,topsep=0.35em]
\item dolma3-short: packs only short documents.
\item dolma3-short + random-long: replaces $34\%$ of the tokens with randomly sampled natural long documents from our candidate pool.
\item dolma3-short + truncLongPPL: replaces the same $34\%$ with documents selected from the same pool using \ours{} at $\tau=0.05$.
\end{itemize}

\begin{figure}[!htbp]
\centering
\includegraphics[width=\linewidth]{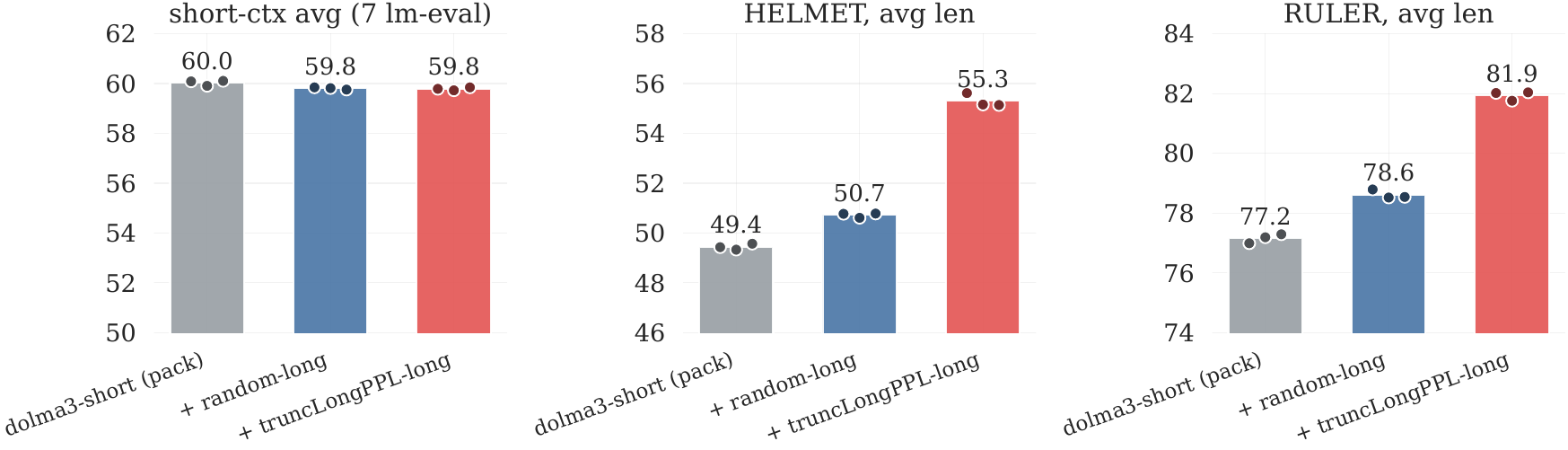}
\caption{Seven-task short-context average and macro-averaged validation HELMET and held-out RULER under a shared 10B-token budget. Bars are means over three independent training seeds and scatter points show the individual seeds. The two 66/34 mixed configurations use the same pool of natural long documents and differ only in random versus \ours{}-based selection. Summary statistics over the three training seeds are reported in Appendix~\ref{app:seed-stability}.}
\label{fig:olmo-production}
\end{figure}

\paragraph{Base-model results.}
Selected long documents raise HELMET from 50.7 to 55.3 ($+4.6$) and RULER from 78.6 to 81.9 ($+3.3$) relative to random-long, while the seven-task short-context average stays at 59.8 (Figure~\ref{fig:olmo-production}). Against dolma3-short, random-long improves HELMET from $49.4$ to $50.7$ ($+1.3$) and RULER from $77.2$ to $78.6$ ($+1.4$); selection gives larger gains of $5.9$ and $4.7$ points. Seven-task short-context averages span 59.8--60.0, with the \ours{} mixture 0.2 point below dolma3-short and close to the base model's 60.1. The long-context advantage holds across the three training seeds (Appendix~\ref{app:seed-stability}); standard deviations around the reported means are at most $0.27$ for all three metrics.

\paragraph{Sensitivity to the mixture ratio.}
Selection improves HELMET by 3.84--4.60 points and RULER by 2.68--3.30 points across the 66/34, 50/50, and 34/66 mixtures. Seven-task short-context averages differ by at most 0.12 point within each random/selected pair (Appendix~\ref{app:mixture-ratios}).

\subsection{Cross-Model Validation at 256K}
\label{sec:cross-model-mixture}
\label{sec:main-qwen3}

We test the mixed recipe on Qwen3-8B Base \citep{yang2025qwen3} and OLMo~3-7B Base \citep{olmo2025olmo} at 256K. Following ProLong \citep{gao2024prolong}, we use 20B continued-training tokens per configuration; the mixed configurations use a 66/34 short/long ratio. We compare dolma3-short, dolma3-short + random-long, and dolma3-short + \ours{}, using its default front-half scoring prefix.

\paragraph{Qwen3-8B.}
\ours{} raises average RULER from 85.61 to 87.08 ($+1.47$) and the 256K endpoint from 73.12 to 77.20 ($+4.08$; Table~\ref{tab:qwen3-main}). The seven-task short-context average is 63.46, close to random-long's 63.42. Compared with dolma3-short, random-long improves the 256K endpoint by 5.73 points, and \ours{} improves it by 9.81 points.

\begin{table}[!htbp]
\caption{Mixed-training validation on Qwen3-8B Base at 256K, with 20B tokens per configuration. Mixed configurations use short:long $=66/34$; dolma3-short uses the same total budget. Short is the seven-task short-context average; RULER is averaged over 8K--256K; RULER 256K is the target-length score. \ours{} uses a front-half scoring prefix (128K). Bold marks the highest value among the displayed configurations in each column.}
\label{tab:qwen3-main}
\centering
\small
\setlength{\tabcolsep}{6pt}
\begin{tabular}{@{}lrrr@{}}
\toprule
Configuration & Short & RULER & RULER 256K \\
\midrule
dolma3-short & 63.42 & 83.72 & 67.39 \\
$+$ random-long & 63.42 & 85.61 & 73.12 \\
$+$ \ours{} & \textbf{63.46} & \textbf{87.08} & \textbf{77.20} \\
\bottomrule
\end{tabular}
\end{table}

\paragraph{OLMo~3-7B.}
\label{sec:transfer}
\label{sec:model-transfer}

\ours{} exceeds random-long by 7.31 average RULER points and 2.88 points at 256K, with a 0.23-point lower seven-task short-context average (Table~\ref{tab:model-transfer}).

\begin{table}[!htbp]
\caption{Cross-model validation on OLMo~3-7B Base at 256K and 20B tokens. Mixed configurations use short:long $=66/34$; dolma3-short uses the same total budget. Metric definitions follow Table~\ref{tab:qwen3-main}. Bold marks the highest value among the displayed configurations in each column.}
\label{tab:model-transfer}
\centering
\small
\setlength{\tabcolsep}{6pt}
\begin{tabular}{@{}lrrr@{}}
\toprule
Configuration & Short & RULER & RULER 256K \\
\midrule
dolma3-short & \textbf{62.11} & 60.60 & 43.73 \\
$+$ random-long & 61.65 & 65.68 & 53.06 \\
$+$ \ours{} & 61.42 & \textbf{72.99} & \textbf{55.94} \\
\bottomrule
\end{tabular}
\end{table}

\FloatBarrier
\subsection{Validation after Supervised Fine-Tuning}
\label{sec:sft}

To test whether selection gains persist after SFT, we fine-tune the three Qwen3-8B continued-training configurations in Table~\ref{tab:qwen3-main} on LongMagpie \citep{gao2025longmagpie}. All three use the same SFT recipe and hyperparameters, with sequences packed to 64K. We then evaluate the fine-tuned models on LongBench v2 \citep{bai2024longbench2} and MRCR \citep{vodrahalli2024michelangelo},\footnote{We use the OpenAI MRCR evaluation set, \url{https://huggingface.co/datasets/openai/mrcr}.} which test long-document reasoning and multi-round retrieval. Complete SFT results and evaluation details appear in Appendix~\ref{app:qwen3-sft-details}.

\begin{table}[!htbp]
\caption{LongBench v2 accuracy (\%) after SFT of Qwen3-8B 256K configurations (means over three evaluation seeds, $N=503$; per-cell standard deviations in Appendix~\ref{app:qwen3-sft-details}). Columns are the overall score and the Short/Medium/Long splits. \ours{} uses a front-half scoring prefix (128K). Best means in each column are in bold.}
\label{tab:longbench-v2}
\centering
\small
\setlength{\tabcolsep}{7pt}
\renewcommand{\arraystretch}{1.2}
\begin{tabular}{@{}l cccc@{}}
\toprule
Configuration & Overall & Short & Medium & Long \\
\midrule
dolma3-short & 33.20 & 42.93 & 28.67 & 25.93 \\
$+$ random-long & 33.87 & \textbf{43.33} & 28.67 & 28.43 \\
$+$ \ours{} & \textbf{35.00} & 42.60 & \textbf{30.53} & \textbf{31.20} \\
\bottomrule
\end{tabular}
\end{table}

\begin{table}[!htbp]
\caption{MRCR accuracy (\%) at context lengths up to 256K after SFT of Qwen3-8B (means over three evaluation seeds, $N=1779$; per-cell standard deviations in Appendix~\ref{app:qwen3-sft-details}). Columns are the overall score and the 2-, 4-, and 8-needle subsets. Best means among the displayed configurations in each column are in bold.}
\label{tab:mrcr}
\centering
\small
\setlength{\tabcolsep}{7pt}
\renewcommand{\arraystretch}{1.2}
\begin{tabular}{@{}l cccc@{}}
\toprule
Configuration & Overall & 2-needle & 4-needle & 8-needle \\
\midrule
dolma3-short & 24.59 & 37.93 & 19.99 & 15.87 \\
$+$ random-long & 24.99 & 38.11 & 20.61 & 16.26 \\
$+$ \ours{} & \textbf{25.61} & \textbf{39.15} & \textbf{20.78} & \textbf{16.90} \\
\bottomrule
\end{tabular}
\end{table}

\paragraph{Results after SFT.}
After SFT, \ours{} scores 35.00 overall on LongBench v2, compared with 33.87 for random-long and 33.20 for dolma3-short (Table~\ref{tab:longbench-v2}). Relative to random-long, its Medium and Long scores improve by 1.87 and 2.77 points, respectively, while its Short score is lower by 0.73 point. On MRCR, \ours{} reaches 25.61, compared with 24.99 for random-long ($+0.62$; Table~\ref{tab:mrcr}), with gains on all three needle subsets. OLMo~3-7B also retains higher overall scores than random-long on both benchmarks after SFT (Appendix~\ref{app:olmo-sft}).

\FloatBarrier
\section{Related Work}
\label{sec:related}

\paragraph{Long-context training recipes.}
Long-context extension is commonly implemented as a late continued-training stage. ProLong emphasizes downstream long-context evaluation rather than perplexity alone \citep{gao2024prolong}; MAI-Thinking-1 finds the preceding midtraining mixture, repacked at 256K, sufficient in its setting \citep{mai2026thinking}; and OLMo~3 combines gzip-filtered scientific PDFs, synthetic augmentation, and a 66/34 short/long mixture \citep{olmo2025olmo}. Foundation-model recipes also differ substantially: Qwen3 trains at 32K with positional and inference-time adaptations \citep{yang2025qwen3}, DeepSeek-V3 extends from 32K to 128K without specifying a document-level selection rule \citep{liu2024deepseek}, and GLM-5 progressively extends to 200K while mixing PPL-filtered natural long documents with synthetic agent and repository data \citep{glm5team2026glm5}. Isolating the effect of document selection from differences in architecture, curriculum, data mixture, and budget requires a controlled comparison. We compare data-construction pipelines under a shared protocol.

\paragraph{Long-document selection.}
Prior selectors use either generic document-level proxies or long-dependency signals. Generic recipes use PPL, deduplication, and length filtering in GLM-5 \citep{glm5team2026glm5} or gzip in OLMo~3, which reports no clear advantage from its evaluated LongPPL-derived alternatives \citep{olmo2025olmo}. Dependency-specific methods measure the predictive value of additional context: LongFilter aggregates token-loss reductions \citep{deng2025longfilter}, Prospector scores dependency strength, distance, and specificity over block pairs \citep{chen2024prospector}, and LongPPL identifies tokens whose likelihood improves with long context \citep{fang2025wrong}. Attention-based LADM \citep{chen2025ladm} and LongAttn \citep{wu2025longattn} instead derive dependency scores from explicit 32K attention statistics, which do not directly scale to our 128K scoring stack. Our comparison covers generic-quality, LLM-judge, and likelihood-based pipelines, uses repacking and random sampling as references, and measures the scoring cost of a reduced-span LongPPL-density pipeline.

\paragraph{Long-context data construction.}
Complementary work constructs long sequences from shorter sources. In-Context Pretraining retrieves and orders semantically related documents \citep{shi2024incontext}; QUEST groups documents sharing predicted queries and keywords \citep{gao2024quest}; NExtLong interleaves meta-chunks with retrieved hard negatives \citep{nextlong2025}; EntropyLong retains retrieved contexts that reduce predictive uncertainty at difficult target positions \citep{jia2025entropylong,jia2026policylong}; and LiteLong synthesizes long-context data resource-efficiently by organizing retrieved documents into coherent long sequences \citep{jia2026litelong}. These methods create explicit cross-document supervision but modify document order or insert retrieved content. We instead select naturally occurring long documents while preserving their within-document structure; combining the two approaches remains future work.

\section{Limitations}
\label{sec:limitations}
The benefits of long-dependency selection vary across capabilities. In our capability analysis, the most consistent gains are in retrieval; QA gains depend on the benchmark, while aggregation and variable tracking show no consistent improvement. These results motivate exploring selection signals that target evidence integration, state tracking, and reasoning over long contexts, and studying how they complement long-range predictability. Our experiments focus on natural long documents from a broad pretraining corpus and exclude repository-level code. Future work could extend the comparison to code and other specialized corpora, examining how long-dependency selection transfers across domains and whether domain-specific selection signals offer further benefits.

\section{Conclusion}
We study natural long-document selection on Qwen2.5-3B at 128K under matched 6B-token budgets. Random sampling of natural long documents outperforms repacking at longer evaluation lengths, and the tested long-dependency pipelines improve on random sampling where the evaluated generic-quality and LLM-judge alternatives do not. LongPPL-density comparisons show that the selected density range and filtering strength matter. \ours{} achieves the highest mean scores across three training seeds, with the clearest gains in retrieval. Front-half scoring retains performance close to full-span scoring with a measured $2.28\times$ speedup in the 128K study. Supplementary experiments support selection at 256K on Qwen2.5-3B and in short/long mixtures across three models. Within these mixtures, selected long data improve long-context scores while keeping seven-task short-context averages close to random-long mixing; Qwen3-8B and OLMo~3-7B retain gains on LongBench v2 and MRCR after SFT. These findings support selecting documents for long-range predictive value while considering filtering strength and scoring cost together.

\bibliography{example_paper}
\bibliographystyle{plainnat}

\appendix

\section{Configurations for the Controlled 128K Pipeline Comparison}
\label{app:pipeline-configurations}

Table~\ref{tab:data-methods} groups the methods into scoring-free constructions and three selector families: generic-quality, LLM-judge, and likelihood-based long-dependency. We use \emph{selection} for the overall process and \emph{selector} for a scoring or ranking method. \textsc{repack} and random-long require no scoring; gzip requires no model inference but computes a document score. The configurations below use Qwen2.5-3B, 128K, and 6B tokens; the mixed-training validation uses the settings in Table~\ref{tab:protocol}. \footnote{We omit attention-based selectors because explicit attention statistics are hard to scale in our 128K scoring stack.}

For the PPL, gzip, and LLM-judge pipelines, we first apply the percentile filters in Table~\ref{tab:data-methods} and then draw a fixed number of qualifying documents by reservoir sampling with seed 42.

\begin{table}[!t]
\caption{Definitions of the data-construction pipelines compared in Section~\ref{sec:selection}, grouped by their selection signals. The first column maps each method to the shorthand used in the result figures. Except for \textsc{repack}, every method operates on the candidate pool $\mathcal{C}$ of natural long documents and preserves document boundaries. All percentile thresholds in this 128K comparison are computed globally over $\mathcal{C}$.}
\label{tab:data-methods}
\centering
\footnotesize
\setlength{\tabcolsep}{3.5pt}
\renewcommand{\arraystretch}{1.12}
\begin{tabular}{@{}L{0.20\linewidth}L{0.32\linewidth}L{0.42\linewidth}@{}}
\toprule
Method & Document signal & Selection rule or construction \\
\midrule
\multicolumn{3}{@{}l}{\textbf{Scoring-free constructions}} \\
\cmidrule(lr){1-3}
\textsc{repack} & None & Pack documents into 128K sequences following the repacking strategy of MAI-Thinking-1; documents may be concatenated. \\
\cmidrule(lr){1-3}
random-long (random in figures) & None & Uniformly sample documents from $\mathcal{C}$ without document scoring. \\
\midrule
\multicolumn{3}{@{}l}{\textbf{Generic-quality selectors}} \\
\cmidrule(lr){1-3}
PPL (top40 / mid40) & Document-level perplexity under Gemma-4-26B-A4B (base) & Select scores above the 60th percentile or between the 30th and 70th percentiles, respectively. \\
\cmidrule(lr){1-3}
gzip (mid60) & gzip compression ratio & Select scores between the 20th and 80th percentiles. \\
\midrule
\multicolumn{3}{@{}l}{\textbf{LLM-judge selector}} \\
\cmidrule(lr){1-3}
LLM-judge (top5) & Fraction of sampled long-distance fragment pairs judged dependent & Select the top 5\% of documents. \\
\midrule
\multicolumn{3}{@{}l}{\textbf{Likelihood-based long-dependency selectors}} \\
\cmidrule(lr){1-3}
LongFilter & Probability-weighted token-loss reduction from short to long context & Select the highest-scoring documents up to the 6B-token budget. \\
\cmidrule(lr){1-3}
Prospector & Dependency strength, distance, and specificity over sampled block pairs & Select the highest-scoring 50\% of documents. \\
\cmidrule(lr){1-3}
LongPPL-density / \ours{} & Density $\rho_W(x)$ of LongPPL key tokens in scoring window $W$ & Select documents satisfying $\rho_W(x)>\tau$; Section~\ref{sec:signals} reports \ours{} ($r=1/2$, $\tau=0.05$), with threshold and span analyses in Section~\ref{sec:robustness} and position analysis in Appendix~\ref{app:window-position}. \\
\bottomrule
\end{tabular}
\end{table}

\subsection{Scoring-Free Constructions: Repacking and Random Sampling}
\textsc{repack} packs documents into 128K training sequences, following the repacking strategy of MAI-Thinking-1 \citep{mai2026thinking}. The random-long pipeline uniformly samples documents from $\mathcal{C}$ without applying a quality or dependency score.

\subsection{Generic-Quality Selectors}
The PPL selector computes document-level perplexity (PPL) using the frozen Gemma-4-26B-A4B (base) scoring model \citep{gemmateam2026gemma4}. The top-$40\%$ rule selects documents above the global $60$th percentile of PPL scores, whereas the middle-$40\%$ rule selects documents between the $30$th and $70$th percentiles. The figure labels top40 ppl and mid40 ppl denote these two rules, respectively.

The gzip selector instead scores each document by the ratio of its gzip-compressed UTF-8 byte length to its original byte length. Lower ratios indicate greater compressibility and can expose highly repetitive documents, whereas unusually high ratios can indicate poorly compressible or noisy content. Following OLMo~3 \citep{olmo2025olmo}, we remove both extremes and select the middle $60\%$ of gzip scores. Appendix~\ref{app:generic-details} gives the exact score definitions.

\subsection{LLM-judge Selector}
The LLM-judge selector (top5 judge in the figures) uses Gemma-4-26B-A4B-instruct \citep{gemmateam2026gemma4} to classify whether a later fragment depends on an earlier fragment across $M$ sampled fragment pairs at varying token distances. We rank documents by the fraction of pairs classified as dependent and select the top $5\%$; Appendix~\ref{app:judge-prompt} gives the exact score definition and the complete user prompt.

\subsection{Likelihood-Based Long-Dependency Selectors}
All three selectors derive their signals from likelihood changes induced by additional context, but differ in granularity and aggregation. Appendix~\ref{app:likelihood-details} gives their score definitions and implementation details.

\paragraph{LongFilter.}
For LongFilter \citep{deng2025longfilter}, we use Gemma-4-26B-A4B (base) \citep{gemmateam2026gemma4} to aggregate continuous token-level loss reductions from short to long context, weighting each reduction by the model's long-context probability of the observed token. We select the highest-scoring documents up to the 6B-token training budget.

\paragraph{Prospector.}
The original Prospector implementation uses OPT-350M as its scoring model \citep{chen2024prospector}. To control the cost of block-pair scoring, our 128K adaptation uses \texttt{SmolLM2-135M} \citep{allal2025smollm2} to score 5,000 sampled pairs of 512-token blocks within each document. Its structured block-pair score combines dependency strength, dependency distance, and dependency specificity; we select the highest-scoring 50\% of documents.

\paragraph{LongPPL-density selection.}
We adapt the LongPPL key-token detector \citep{fang2025wrong} to document selection. A token is marked as key when a long prefix improves its log-likelihood over a 4K short prefix by more than $\alpha$ and its long-context log-likelihood exceeds $\beta$. For a scoring window, let $W$ contain its valid target positions after excluding the document's first 4K tokens. We select documents whose key-token density $\rho_W(x)=|W|^{-1}\sum_{i\in W}k_i$ exceeds $\tau$; Equation~\ref{eq:prefix-density} gives the prefix case. In the Qwen2.5-3B 128K sensitivity study, we fix the scoring model, tokenizer, short-context construction, and $(\alpha,\beta)$, varying only $W$ and $\tau$. Section~\ref{sec:robustness} evaluates thresholds and prefix fractions from $r=1/8$ to $r=1$ at $L=128$K. Appendices~\ref{app:window-position} and~\ref{app:scattered} compare window positions at matched nominal spans and more complex layouts.

\section{Implementation Details for Generic-Quality Selectors}
\label{app:generic-details}

For a document $x$, the PPL selector uses the frozen scoring model $p_{\theta_0}$ and computes
\begin{equation}
    s_{\mathrm{PPL}}(x)=\exp\left(-\frac{1}{|x|}\sum_{i=1}^{|x|}\log p_{\theta_0}(x_i\mid x_{<i})\right).
\end{equation}
The gzip selector uses the ratio of level-6 gzip-compressed to original UTF-8 byte length:
\begin{equation}
    s_{\mathrm{gzip}}(x)=\frac{|\operatorname{gzip}_6(\operatorname{UTF8}(x))|}{|\operatorname{UTF8}(x)|}.
\end{equation}

\section{LLM-judge Selector Details and Prompt}
\label{app:judge-prompt}

The LLM-judge pipeline (Section~\ref{sec:data-methods}) tests whether a later fragment depends on an earlier one. For $M$ sampled pairs with predictions $y_m$, the document score is the fraction judged dependent:
\begin{equation}
    s_{\mathrm{judge}}(x)=\frac{1}{M}\sum_{m=1}^{M}\mathbb{I}[y_m=\text{dependent}].
\end{equation}
The judge is queried with a single user message and no separate system prompt; the complete message is reproduced verbatim in the box below. The placeholders \texttt{\{head\_text\}} and \texttt{\{tail\_text\}} are replaced by the earlier and later fragment, respectively (each truncated to its leading 512 and trailing 256 tokens, with the middle elided as ``\ldots'' when longer). The model is instructed to return a single JSON object with a one-sentence rationale, a binary \texttt{verdict}, and the matched criterion.

\begin{tcolorbox}[
  breakable,
  colback=takeawaybody,
  colframe=takeawayborder,
  colbacktitle=takeawaytitle,
  coltitle=black,
  fonttitle=\bfseries,
  title={Prompt used by the LLM-judge selector},
  boxrule=0.7pt,
  arc=1.5mm,
  left=6pt,
  right=6pt,
  top=5pt,
  bottom=5pt
]
\begin{Verbatim}[fontsize=\scriptsize,breaklines=true,breakanywhere=true]
You are a rigorous expert in judging inter-passage DEPENDENCY within a long
document.

Below are TWO NON-ADJACENT fragments from the SAME long document: Fragment 1
comes earlier and Fragment 2 comes later, and they may be far apart in the
original text. To save space, an overly long fragment keeps only its beginning
and end, with the middle elided as "...".

[Fragment 1 (earlier)]
{head_text}

[Fragment 2 (later)]
{tail_text}

# Task
Imagine a reader who has read ONLY Fragment 2 and has never seen Fragment 1:
would they fail to correctly understand Fragment 2 because some specific piece
of information supplied by Fragment 1 is missing?
In other words - does Fragment 2 contain a "dangling" reference / premise /
constraint / response whose gap can ONLY be filled by Fragment 1?

# Criteria for "DEPENDENT (1)" (any one suffices; the dependency MUST point to
specific content in Fragment 1)
  C1 Entity reference
    Meaning: Fragment 1 first defines or introduces a specific entity, person,
    concept, term, abbreviation, symbol, variable, or label; Fragment 2 then
    uses it directly without restating its identity or meaning.
    Typical signs: Fragment 2 says "this method / the above system / he / it /
    this parameter / as mentioned earlier", or directly uses a proper noun,
    abbreviation, or label that only Fragment 1 explains.
    Positive: F1 defines "we denote our model as M"; F2 says "M reaches 90% on
    the test set" - a reader who skips F1 has no idea what M is.
    Boundary: if the entity is common domain knowledge (e.g. "CPU", "DNA") or
    Fragment 2 explains it itself, it is NOT a dependency.

  C2 Instruction-response
    Meaning: Fragment 1 poses a specific question, task, requirement, prompt,
    or to-do; Fragment 2 is precisely its answer, solution, execution, or
    fulfillment.
    Typical signs: Fragment 1 is a question / problem statement / requirement,
    and Fragment 2 is the answer / solution / implementation / conclusion; or
    Fragment 2 continues with "to answer the above / to accomplish that task".
    Positive: F1 asks "how to reduce GPU memory usage?"; F2 answers "using a
    paged KV cache cuts it by 40%".
    Boundary: if Fragment 2 merely happens to discuss a related topic but does
    not address that specific question in Fragment 1, it does NOT count.

  C3 Numeric / constraint carry-over
    Meaning: Fragment 1 sets a specific value, parameter, unit, range, initial
    condition, assumption, or constraint; Fragment 2 builds on it to continue
    computing, deriving, or arguing.
    Typical signs: Fragment 2 uses "that value / the above parameter / under
    this setting", or directly reuses a constant, threshold, formula symbol, or
    experimental config from Fragment 1 without re-declaring it.
    Positive: F1 sets "learning rate = 1e-4, batch = 256"; F2 says "under this
    config it converges in 3 epochs".
    Boundary: if the two fragments each provide their own independent values and
    do not reuse each other's, it does NOT count.

  C4 Causal / temporal continuation
    Meaning: Fragment 2 is the direct result, subsequent step, next stage, or
    temporal continuation of an event, action, reasoning, or procedure
    described in Fragment 1, forming a before->after / cause->effect chain.
    Typical signs: Fragment 2 continues Fragment 1's narrative or reasoning with
    "therefore / thus / then / next / after that / finally"; or it is clearly
    the next step of the same procedure or storyline.
    Positive: F1 "mix the reactants and heat to 80C"; F2 "once the solution
    turns clear, filter it" - the next step of the same procedure.
    Boundary: if they are merely two unrelated facts under the same topic with
    no causal or sequential link, it does NOT count.

  C5 Citation / call-back
    Meaning: Fragment 2 explicitly cites, responds to, rebuts, supplements, or
    extends a specific claim, stance, conclusion, evidence, or viewpoint
    established in Fragment 1.
    Typical signs: Fragment 2 contains call-backs like "as noted above / as
    argued earlier / on the contrary / building on this / this confirms the
    earlier conclusion", and the referent is specific content in Fragment 1.
    Positive: F1 argues "a long context does not necessarily lower perplexity";
    F2 "based on the above conclusion, we redesigned the evaluation".
    Boundary: if Fragment 2 only vaguely mentions a similar viewpoint rather
    than calling back the specific claim/evidence built in Fragment 1, it does
    NOT count.

# When to judge "NOT DEPENDENT (0)" (avoid false positives)
  - Fragment 2 is self-contained and can be fully understood without Fragment 1.
  - The two fragments only share "same topic / same domain / similar wording /
    both mention some common-knowledge entity", but Fragment 2 does not actually
    need any specific information from Fragment 1 - such surface similarity does
    NOT count as dependency.
  - The two fragments are independent, parallel items/sections (e.g.
    encyclopedic listings, mutually-unreferenced problem lists), even if the
    same terminology recurs.

# Output
First give a one-sentence rationale that pinpoints WHICH specific element of
Fragment 2 does or does not depend on Fragment 1 (be concrete, not generic),
then give the verdict. Output ONLY the JSON, nothing else:
{"reasoning":"...","verdict":0 or 1,"criterion":"C1/C2/C3/C4/C5 or none"}
\end{Verbatim}
\end{tcolorbox}

\section{Implementation Details for Likelihood-Based Selectors}
\label{app:likelihood-details}

\paragraph{LongFilter.}
LongFilter \citep{deng2025longfilter} contrasts token losses under long and short contexts. Let $\ell_i^{\mathrm{long}}$ and $\ell_i^{\mathrm{short}}$ denote the corresponding negative log-likelihoods. Its document score is
\begin{equation}
    s_{\mathrm{LongFilter}}(x)=\frac{1}{|x|}\sum_{i=1}^{|x|}
    \exp(-\ell_i^{\mathrm{long}})
    \bigl(\ell_i^{\mathrm{short}}-\ell_i^{\mathrm{long}}\bigr),
\end{equation}
which weights each loss reduction by the model's long-context probability of the observed token.

\paragraph{Prospector.}
The original Prospector pipeline uses OPT-350M and divides 32K documents into $N=256$ blocks of 128 tokens \citep{chen2024prospector}. Our 128K adaptation uses the \texttt{SmolLM2-135M} scoring model and keeps $N=256$ by increasing the block size to 512 tokens. For an earlier block $c_j$ and a later block $c_i$, it computes dependency strength
\begin{equation}
    \operatorname{DST}(i,j)=1-\frac{\operatorname{PPL}(c_i\mid c_j)}{\operatorname{PPL}(c_i)},
\end{equation}
and dependency distance $\operatorname{DDI}(i,j)=(i-j)/(N-1)$. With $\Delta_{ij}=\operatorname{PPL}(c_i)-\operatorname{PPL}(c_i\mid c_j)$ and $q_{ij}=\exp(\Delta_{ij})/\sum_{k<i}\exp(\Delta_{ik})$, dependency specificity for target blocks with at least two possible predecessors ($i\geq 3$) is the normalized inverse entropy
\begin{equation}
    \operatorname{DSP}(i)=1-\frac{-\sum_{j<i}q_{ij}\log q_{ij}}{\log(i-1)}.
\end{equation}
The document score is the sum of $(a\operatorname{DST}(i,j)+b\operatorname{DDI}(i,j))\operatorname{DSP}(i)$ over 5,000 sampled block pairs with $i\geq 3$ satisfying $\operatorname{DST}(i,j)>0.1$. We follow \citet{chen2024prospector} with $(a,b)=(1,1)$.

\paragraph{LongPPL key-token density.}
Let $x_{<i}^{\mathrm{long}}$ be the available long prefix of target token $x_i$, and $x_{<i}^{\mathrm{short}}$ its most recent $K$ prefix tokens. Following LongPPL \citep{fang2025wrong}, we define the long--short likelihood difference as
\begin{equation}
    \operatorname{LSD}(i)=\log p_{\theta_0}(x_i\mid x_{<i}^{\mathrm{long}})
    -\log p_{\theta_0}(x_i\mid x_{<i}^{\mathrm{short}})
\end{equation}
and let $\operatorname{LCL}(i)=\log p_{\theta_0}(x_i\mid x_{<i}^{\mathrm{long}})$. A token is classified as a key token when
\begin{equation}
    k_i=\mathbb{I}[\operatorname{LSD}(i)>\alpha
    \;\wedge\;\operatorname{LCL}(i)>\beta].
\end{equation}
The LCL condition removes tokens that remain intrinsically unpredictable even with long context. The first $K=4096$ tokens of each document provide initial context and contribute to neither the numerator nor the denominator of the density. For any scoring window, let $W$ denote its valid target positions after excluding these initial tokens. The document-level key-token density is
\begin{equation}
    \rho_W(x)=\frac{1}{|W|}\sum_{i\in W} k_i,
\end{equation}
and the corresponding selected set is
\begin{equation}
    \mathcal{D}_{W,\tau}=\{x\in\mathcal{C}:\rho_W(x)>\tau\}.
\end{equation}
Thus, the document thresholds from 0.01 to 0.10 in our experiments apply to $\rho_W$, not to the LongPPL token-level parameters $\alpha$ and $\beta$. For \ours{} with prefix fraction $r$, let $T_r=\lfloor rL\rfloor>K$, where $L$ is the nominal target-context reference span and $r$ specifies the prefix fraction. As in Equation~\ref{eq:prefix-density},
\begin{equation}
    W_r=\{K+1,\ldots,T_r\},\qquad |W_r|=T_r-K.
\end{equation}
For example, a 16K prefix uses positions $(4\mathrm{K},16\mathrm{K}]$ and a density denominator of 12K tokens. The default is $r=1/2$; $r=1$ is the full-span reference. The prefix endpoint and valid target positions are counted after tokenization with the scoring model's tokenizer. The fraction is relative to $L$, rather than to each raw document's variable length.

\paragraph{Scoring configuration and windows.}
For the controlled Qwen2.5-3B 128K study, the frozen scoring model $p_{\theta_0}$ is Gemma-4-26B-A4B (base) \citep{gemmateam2026gemma4}, served in bf16 through vLLM \citep{kwon2023vllm} with a FlashAttention backend \citep{dao2022flashattention} on H20 GPUs. We use its native tokenizer, set the short-context length to $K=4096$, and follow LongPPL \citep{fang2025wrong} by fixing the token-level thresholds to $(\alpha,\beta)=(2.0,-2.0)$. All LongPPL scoring-window lengths and the 128K truncation length are measured after tokenization with this Gemma tokenizer. We compute LSD and LCL for every valid target token beyond the initial $K$ positions from full-context log probabilities (\texttt{echo}, one log probability per position), without target-token subsampling; short-context losses are computed using sliding windows with a stride of $1024$ tokens. We score documents up to 128K Gemma tokens. Full-span scoring of the complete pool requires on the order of $10^4$--$10^5$ H20 GPU-hours, making the scoring span an important computational variable.

At $L=128$K, we score prefixes with $r\in\{1/8,1/4,1/2,1\}$, corresponding to prefix endpoints of 16K, 32K, 64K, and 128K; tail windows with fractions $1/4$ or $1/2$; or the intersection of documents independently selected by front-half and tail-half windows. At $L=256$K, the tested prefix fractions $r=1/4$ and $r=1/2$ correspond to prefix endpoints of 64K and 128K, respectively. Across the Qwen2.5-3B 128K document-selection sweeps, we hold the scoring model, tokenizer, short-context construction, and token-level thresholds $(\alpha,\beta)$ fixed; only the document-level threshold $\tau$ and scoring window $W$ vary.

\section{OLMo-Style Middle-60\% Selection Control}
\label{app:olmo-reconcile}

OLMo~3 selects scientific PDFs by removing the most and least compressible 20\% and reports no clear advantage over gzip for its LongPPL-derived alternatives based on key-token fraction or positional spread \citep{olmo2025olmo}. Its evaluated LongPPL rules include one-sided bottom-20\% removal and symmetric top-and-bottom-20\% removal.

To examine this selection convention, we reproduce the symmetric middle-60\% LongPPL-density-versus-gzip comparison in our setting. Table~\ref{tab:olmo-reconcile} shows the same qualitative result as OLMo~3: the two pipelines differ by at most 1.17 points across all evaluated budgets and by at most 0.51 points from 4B tokens onward. At the shared 6B-token budget, both remain below the random-long baseline of 51.66 validation HELMET / 78.05 held-out RULER. LongPPL-density selection consistently yields the higher held-out RULER score, whereas gzip yields the higher validation HELMET score from 6B tokens onward; under this design, these small differences do not establish a reliable ordering. We match the symmetric middle-range selection component of OLMo~3 within our own training recipe.

\begin{table}[!htbp]
\caption{OLMo-style middle-60\% selection comparison in our training setup across budgets of 2B--10B tokens. Columns are grouped into validation HELMET and held-out RULER; within each group, the two subcolumns compare LongPPL-density and gzip selection. The higher value in each (budget, metric) pair is bold.}
\label{tab:olmo-reconcile}
\centering
\small
\begin{tabular}{lcccc}
\toprule
& \multicolumn{2}{c}{HELMET} & \multicolumn{2}{c}{RULER} \\
\cmidrule(lr){2-3}\cmidrule(lr){4-5}
Training tokens & LongPPL-density & gzip & LongPPL-density & gzip \\
\midrule
2B  & \best{50.41} & 49.24 & \best{76.67} & 75.86 \\
4B  & \best{50.56} & 50.36 & \best{77.84} & 77.55 \\
6B  & 50.40 & \best{50.71} & \best{77.91} & 77.45 \\
8B & 50.33 & \best{50.68} & \best{77.93} & 77.72 \\
10B & 50.54 & \best{51.05} & \best{78.15} & 78.01 \\
\bottomrule
\end{tabular}
\end{table}

The LongPPL-density rules evaluated by OLMo~3 included removing the bottom 20\% of documents and removing both the top and bottom 20\%. The former retains high-density documents within a broad candidate set, whereas the latter excludes the highest-density tail. In our controlled Qwen2.5-3B study at the shared 6B-token budget, full-span selection of the high-key-token-density tail yields better downstream performance than retaining the middle 60\% of documents. Our density-threshold configurations apply the absolute key-token-density criterion $\rho_W(x)>\tau$ with $\tau\in[0.03,0.05]$. With full-span scoring ($r=1$, $L=128$K), these thresholds select 216{,}448 (7.2\%) and 51{,}959 (1.7\%) documents, respectively, from the 3M-document candidate pool. Together with the threshold sweep in Section~\ref{sec:threshold-robustness}, these results suggest that the selected density range and strength of filtering affect the downstream benefit of LongPPL-based selection in our setting. Differences in these choices may help explain the differing findings across studies.

\section{Front-Half versus Full-Span Scoring at 128K: Agreement Diagnostics and Cost}
\label{app:agreement-cost}

At $L=128$K, we compare front-half ($r=1/2$, a 64K prefix) and full-span ($r=1$, a 128K prefix) scoring on the subset of documents that exceed a low density threshold in at least one window. Their per-document key-token densities are strongly correlated (Pearson $r=0.83$; Spearman $\rho=0.77$). This selection-conditioned audit measures association within the subset, not estimator accuracy over the full candidate pool. Let $S_{1/2}(\tau)$ and $S_1(\tau)$ denote the sets selected by front-half and full-span scoring at threshold $\tau$. Across $\tau\in\{0.02,0.03,0.05\}$, the one-way containment rate from front-half to full-span selection
\begin{equation}
    \frac{|S_{1/2}(\tau)\cap S_1(\tau)|}{|S_{1/2}(\tau)|}
\end{equation}
ranges from $76\%$ to $92\%$, with full-span scoring generally selecting a larger set. This quantity is the fraction of front-half-selected documents also selected by full-span scoring, not a symmetric agreement measure.

For this model at $L=128$K, front-half scoring ($r=1/2$, 64K) rather than full-span scoring ($r=1$, 128K) reduces the estimated prefill FLOPs to approximately $0.37\times$ (a $2.7\times$ reduction), based on its 25 sliding-window and 5 full-attention layers. We also directly measure per-document wall-clock time for the complete scoring pipeline. The full-context loss (stage~1) decreases from $7.5$ to $2.55$\,s ($2.94\times$), and the sliding-window short-context loss (stage~2) decreases from $9.05$ to $4.70$\,s ($1.93\times$). The combined time therefore decreases from $16.55$ to $7.25$\,s per document, yielding an end-to-end $2.28\times$ speedup.

The stages scale differently. Stage~1 comprises a single long-sequence prefill whose attention incurs a superlinear $O(n^2)$ term, so halving the sequence length reduces its cost by more than $2\times$. Stage~2 computes short-context losses using sliding windows with a fixed stride of $1024$ tokens. The number of windows scales linearly with document length, so halving the scoring span approximately halves the number of windows. Because stage~2 accounts for most of the total cost ($55\%$ at 128K and $65\%$ at 64K), the end-to-end $2.28\times$ speedup is closer to the stage-2 ratio of approximately $2\times$ than to the stage-1 or FLOPs ratio.

\section{Additional LongPPL-Density Analyses}
\label{app:window-robustness}

\subsection{Position-Wise Key-Token Density}
\label{app:keytoken-position}

\begin{figure}[!tb]
\centering
\includegraphics[width=0.60\linewidth]{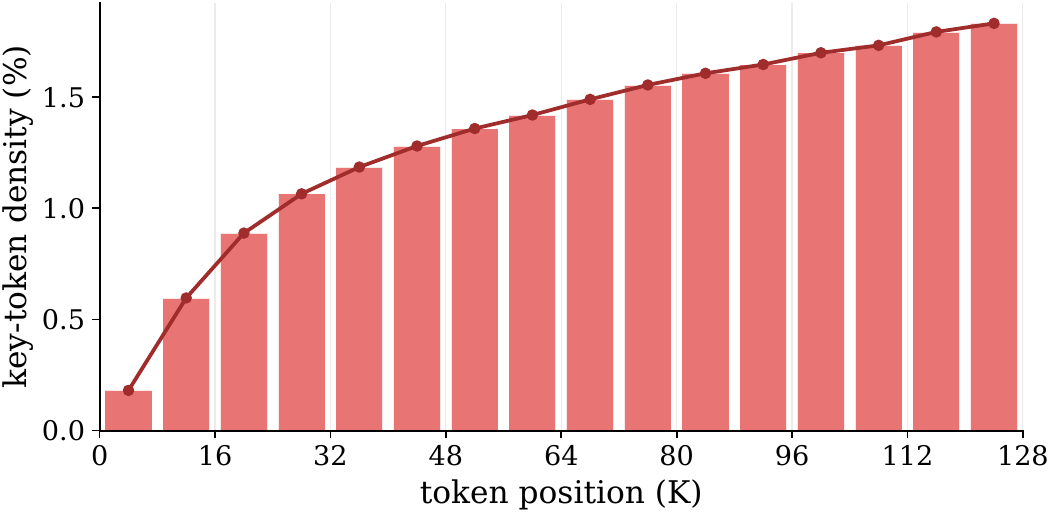}
\caption{Raw mean LongPPL key-token density across 8K positional bins for 30{,}000 natural long documents truncated to 128K tokens. Density increases with position, but the available prefix also grows; the curve is therefore descriptive rather than evidence of a position-only effect.}
\label{fig:keytoken-position}
\end{figure}

Figure~\ref{fig:keytoken-position} shows that the raw mean density increases monotonically across the 8K bins. Later target tokens have access to a longer prefix, while early targets have less context beyond the 4K short-context reference. The curve therefore reflects both document structure and the growing available prefix. The downstream comparison in Section~\ref{sec:window-length} evaluates the front-half window in terms of scoring cost and quality.

\subsection{Sensitivity to the Training-Token Budget}
\label{app:budget-sensitivity}

The controlled Qwen2.5-3B 128K comparisons fix the training budget at 6B tokens. We vary the budget from 2B to 10B tokens while holding the front-half ($r=1/2$) scoring window fixed to test whether the threshold ordering persists across continued-training scales.

\begin{figure}[!tb]
\centering
\includegraphics[width=0.80\linewidth]{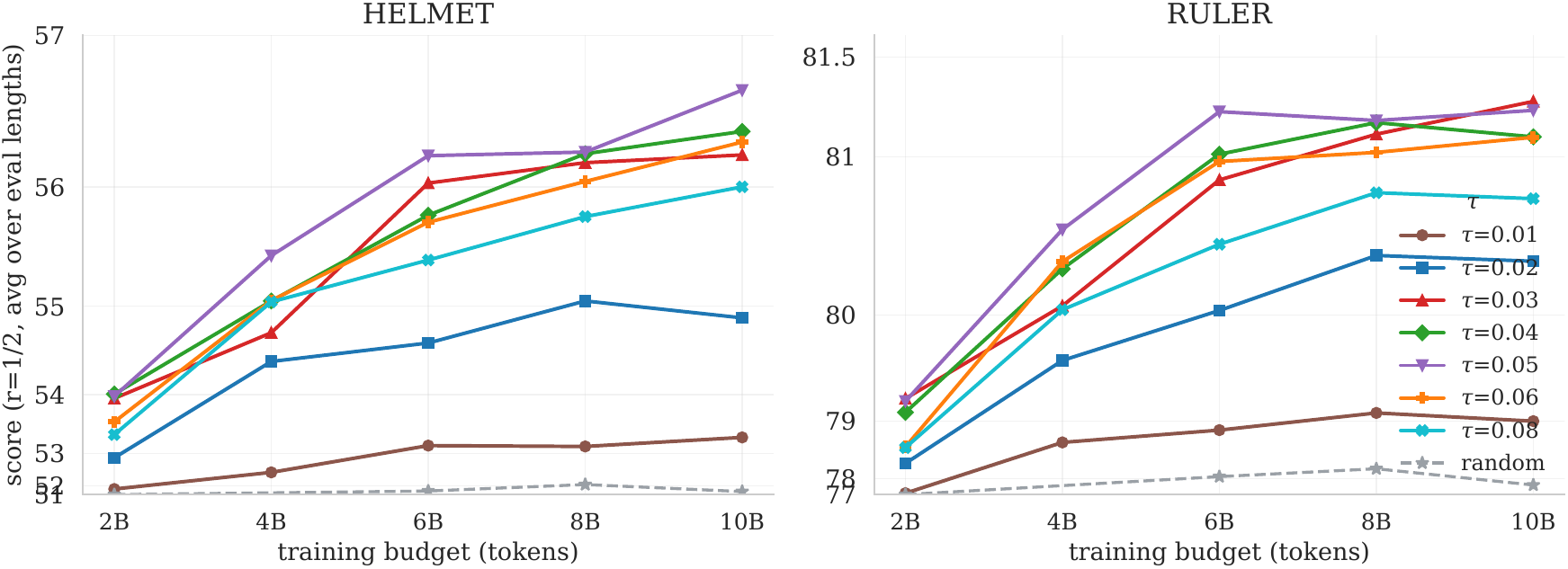}
\caption{Validation HELMET (left) and held-out RULER (right) by continued-training budget for front-half ($r=1/2$) LongPPL-density selection at $L=128$K. Low thresholds remain lower, while $\tau\in\{0.03,0.04,0.05,0.06\}$ forms a higher cluster from approximately 6B tokens onward.}
\label{fig:budget-front64}
\end{figure}

As shown by HELMET validation in Figure~\ref{fig:budget-front64}, the permissive $\tau\in\{0.01,0.02\}$ configurations remain lower, indicating that additional training tokens do not compensate for insufficient selectivity. Thresholds from 0.03 to 0.06 form a higher-performing cluster from approximately 6B tokens onward, with generally small subsequent gains. Within this cluster, $\tau=0.06$ has no consistent validation advantage over $\tau=0.05$, while $\tau=0.08$ remains lower. We therefore retain $\tau=0.05$ as the default high-density configuration.

\subsection{Window Position at Threshold 0.05}
\label{app:window-position}

Having fixed the nominal window length, we ask \emph{where} in the document the density should be measured. At $L=128$K, we compare front-half and tail-half windows, the full span, and eight scattered windows totaling half of the reference span (each 8K). We also evaluate the intersection of the document sets independently selected by the front-half and tail-half windows.

\begin{figure}[!tb]
\centering
\includegraphics[width=0.80\linewidth]{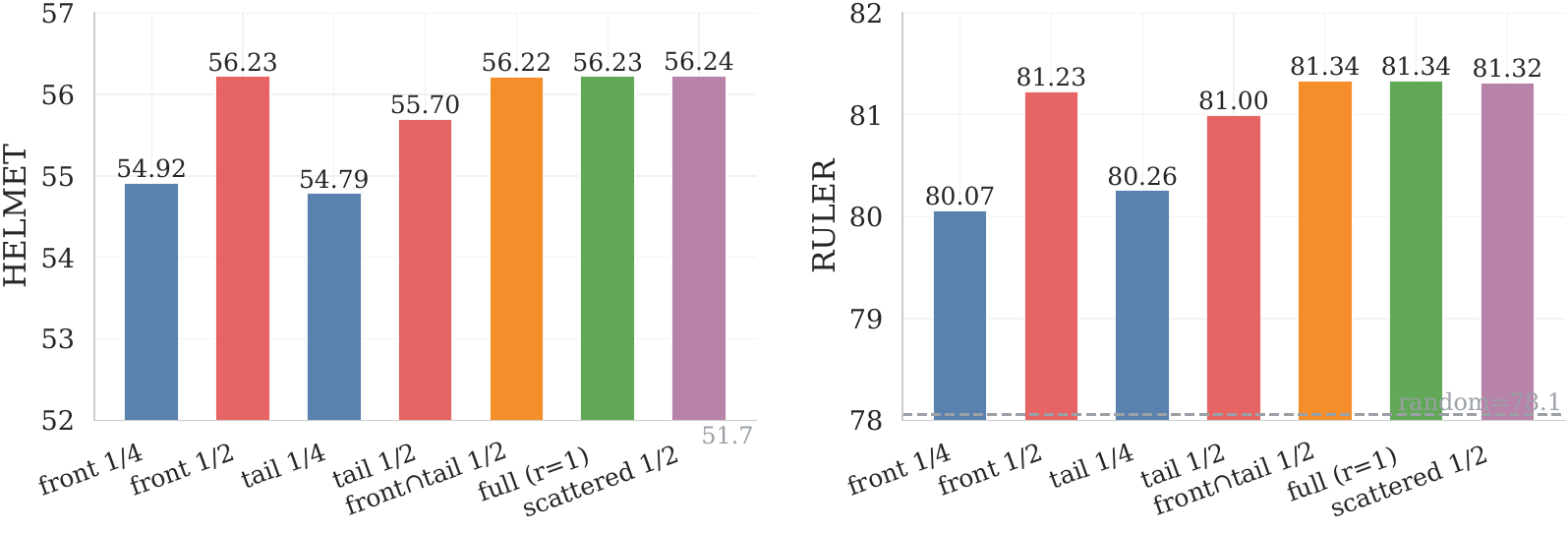}
\caption{Macro-averaged validation HELMET (left) and held-out RULER (right) for scoring-window layouts at $\tau=0.05$ under the 6B-token budget. Dashed lines mark random sampling of natural long documents. All layouts exceed this baseline. The front$\cap$tail label denotes the intersection of the document sets selected by the front-half and tail-half windows.}
\label{fig:window-position}
\end{figure}

Figure~\ref{fig:window-position} shows small differences among the tested positions. On HELMET validation, the front-half window achieves 56.23 compared with 55.70 for the tail-half window, and neither the intersection of the two selected document sets nor the scattered layout provides a consistent validation advantage over the front-half scoring window. We select the front-half scoring window as the simplest contiguous default. With this default fixed, held-out RULER scores are 81.23 for the front-half window and 81.00 for the tail-half window.

\subsection{Scattered versus Contiguous Windows}
\label{app:scattered}

To test whether broader positional coverage helps, Table~\ref{tab:scattered} compares eight randomly located windows with a total nominal span of $L/2$ (eight 8K windows at $L=128$K) with one contiguous front-half window at the same threshold. Across $\tau\in\{0.03,0.04,0.05\}$, validation HELMET differs by approximately one point or less, with no consistent direction. We therefore keep the front-half default. Held-out RULER, evaluated after fixing this choice, also differs by less than one point with no consistent direction.

\begin{table}[!htbp]
\caption{Scattered ($8\times L/16$, randomly located) versus contiguous (front-half) scoring at shared absolute thresholds, $L=128$K and 6B tokens. Both layouts have a nominal span of $L/2$; densities are normalized by their valid target counts as defined in Appendix~\ref{app:likelihood-details}. Cells give validation HELMET / held-out RULER averaged over evaluation lengths.}
\label{tab:scattered}
\centering
\small
\begin{tabular}{lcc}
\toprule
Threshold $\tau$ & Scattered $8\times L/16$ & Contiguous front-half \\
\midrule
0.03 & 55.01 / 80.55 & 56.03 / 80.87 \\
0.04 & 56.23 / 80.97 & 55.79 / 81.01 \\
0.05 & 56.24 / 81.32 & 56.23 / 81.23 \\
\bottomrule
\end{tabular}
\end{table}

\subsection{Window-Length Sensitivity at Threshold 0.03}
\label{app:tau03}

The Qwen2.5-3B window-length analysis uses the default $\tau=0.05$, the least restrictive point on the $0.05$--$0.08$ validation plateau in Section~\ref{sec:robustness}. We repeat that comparison at the more permissive threshold $\tau=0.03$. The same pattern holds as at $\tau=0.05$: the $r=1/2$ and $r=1$ pipelines have similar scores, whereas $r=1/8$ is lower at long evaluation lengths.

\begin{figure}[!htbp]
\centering
\includegraphics[width=0.80\linewidth]{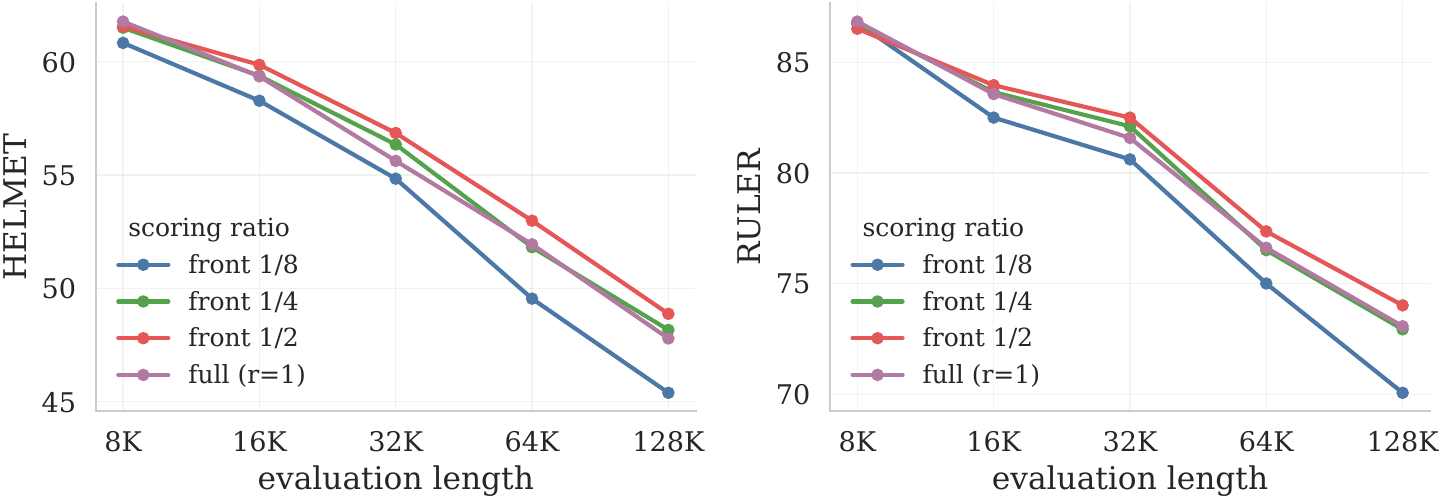}
\caption{Scores by evaluation length for prefix fractions $r\in\{1/8,1/4,1/2,1\}$ ($L=128$K, $\tau=0.03$, 6B training tokens). At 128K evaluation, $r=1/8$ scores lower, while $r=1/2$ and $r=1$ remain close (Figure~\ref{fig:window-eval-length}).}
\label{fig:window-eval-length-003}
\end{figure}

\FloatBarrier
\subsection{Scoring Windows for Length Extension}
\label{app:length-window}

The default \ours{} configuration uses $r=1/2$, giving scoring prefixes of 64K and 128K for target contexts of 128K and 256K, respectively. At 256K, we compare this with a 64K prefix ($r=1/4$), which retains the absolute scoring length used by the 128K default. This comparison examines whether to retain that cutoff or extend the scoring prefix with the target context. At 128K, $r=1/4$ gives a 32K prefix.

\begin{table}[!htbp]
\caption{Scoring-window comparison on Qwen2.5-3B with 6B training tokens per configuration. The 128K average covers 8K--128K, and the 256K average covers 8K--256K. In the 128K column, $\dagger$ denotes the reference-run result from the $\tau=0.05$ window ablation in Figure~\ref{fig:window-eval-length}; the default configuration reports mean $\pm$ standard deviation over three training seeds. The 256K columns retain the reported point estimates. Bold marks the higher score in each column.}
\label{tab:qwen25-window-transfer}
\centering
\small
\setlength{\tabcolsep}{5pt}
\begin{tabular}{@{}lcrr@{}}
\toprule
& 128K training & \multicolumn{2}{c}{256K training} \\
\cmidrule(lr){2-2}\cmidrule(lr){3-4}
Scoring configuration & RULER & RULER & RULER 256K \\
\midrule
\ours{} ($r=1/4$) & $80.07^{\dagger}$ & 78.01 & 66.41 \\
\ours{} ($r=1/2$, default) & $\mathbf{81.31\pm0.14}$ & \textbf{78.63} & \textbf{67.05} \\
\bottomrule
\end{tabular}
\end{table}

At 256K, the 64K prefix exceeds random-long and LongFilter by 2.20 and 0.76 average RULER points. Extending the scoring prefix to 128K adds 0.62 point on the average and 0.64 point at the endpoint. These results support using the 128K scoring prefix in this 256K setting.

\FloatBarrier
\section{Coarse Domain Reweighting Control}
\label{app:proportional}

\begin{figure}[!htbp]
\centering
\includegraphics[width=0.8\linewidth]{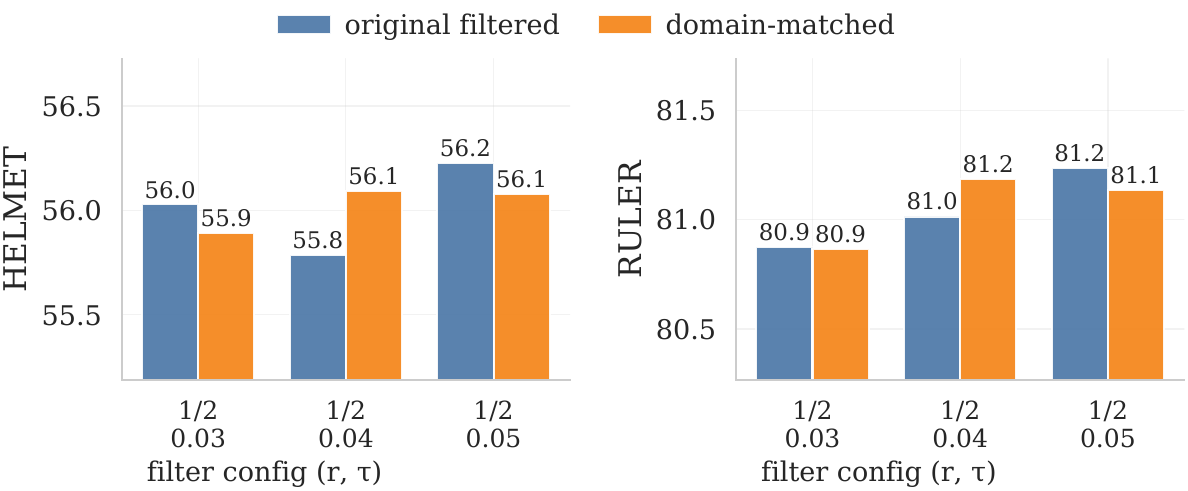}
\caption{Domain-mixture control for \ours{} at $L=128$K under the 6B-token budget. Resampling restores the candidate pool's coarse domain proportions and changes validation HELMET (left) by at most 0.3 point and held-out RULER (right) by at most 0.2 point.}
\label{fig:proportional}
\end{figure}

Selection may change the proportions of books, web pages, and papers. We assess this shift by resampling documents selected by \ours{} to match the domain proportions of the original candidate pool and comparing performance before and after resampling. Across all three thresholds, domain-matched resampling changes HELMET by at most 0.3 point and RULER by at most 0.2 point, with no systematic direction. The gains persist after matching coarse domain proportions.

\section{Short-Context Performance under Long-Context Continued Training}
\label{app:short-performance}

\paragraph{Short-data source.}
\label{app:short-data}
For the short-data component of the mixed-training experiments, we use the midtraining data of OLMo~3-7B \citep{olmo2025olmo}. We denote the short-only packed configuration as dolma3-short.

\begin{figure}[!htbp]
\centering
\includegraphics[width=\linewidth]{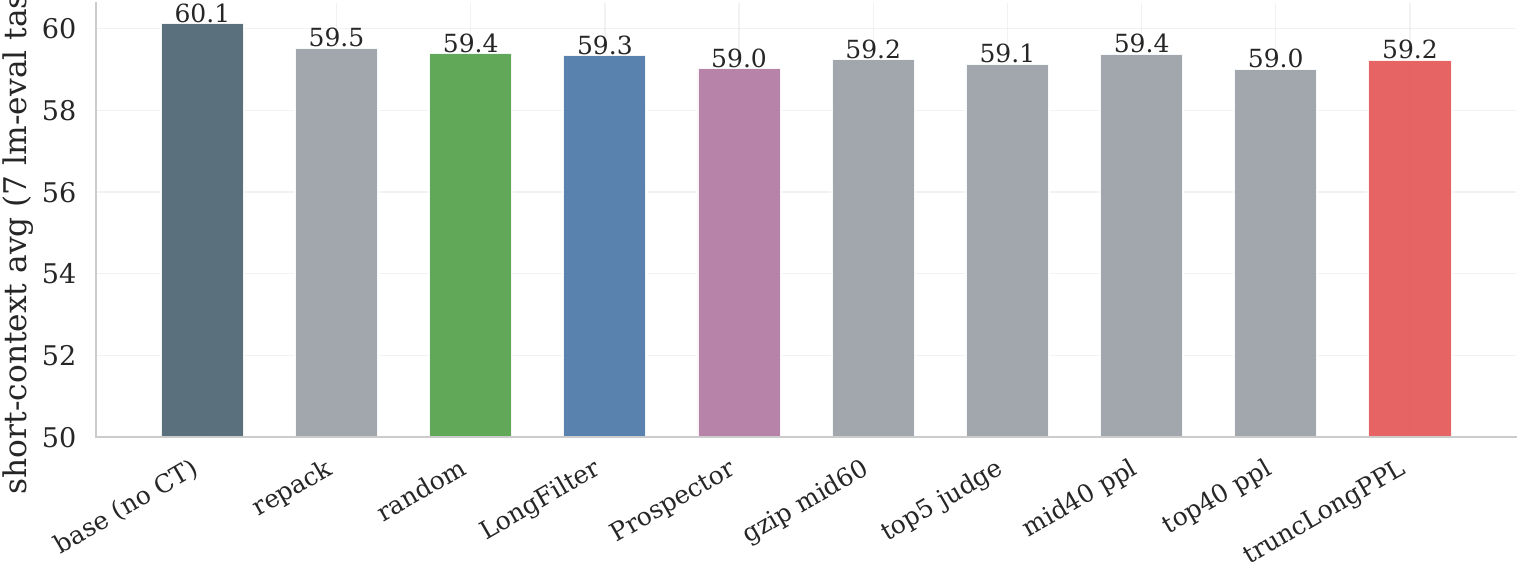}
\caption{Seven-task short-context averages before and after 6B-token continued training. The base model scores 60.1; continued models score 59.0--59.5, including 59.2 for \ours{}.}
\label{fig:short-performance}
\end{figure}

\begin{table}[!htbp]
\caption{Per-task short-context accuracy (\%) for the uncontinued base model (\textsc{Base}, no long-context extension) and the three fixed-mixture configurations in Section~\ref{sec:mixture-qwen25} under the shared 10B-token budget: dolma3-short, and the 66/34 mixtures adding random versus \ours{}-selected long documents. Values are for one representative training seed; Table~\ref{tab:seed-stability} reports averages over three training seeds for the final column. Columns are the seven \texttt{lm-evaluation-harness} tasks and their unweighted average. The highest value in each column is bold.}
\label{tab:short-pertask}
\centering
\small
\setlength{\tabcolsep}{4pt}
\renewcommand{\arraystretch}{1.12}
\begin{tabular}{@{}l ccccccc c@{}}
\toprule
Configuration & ARC-C & ARC-E & HellaSwag & LAMBADA & LogiQA & PIQA & WinoG. & Avg. \\
\midrule
\textsc{Base} (no CT) & 45.05 & 77.36 & \textbf{55.00} & \textbf{66.89} & \textbf{29.95} & \textbf{78.24} & 68.35 & \textbf{60.12} \\
\midrule
dolma3-short & \textbf{48.21} & \textbf{79.17} & 54.27 & 66.31 & 26.27 & 78.18 & 68.19 & 60.09 \\
$+$ random-long & 47.01 & 78.58 & 54.17 & 65.63 & 27.34 & 78.02 & 68.19 & 59.85 \\
$+$ \ours{} & 46.84 & 78.45 & 54.10 & 65.26 & 27.34 & 78.13 & \textbf{68.43} & 59.79 \\
\bottomrule
\end{tabular}
\end{table}

\begin{table}[!htb]
\caption{Seed stability of the three fixed-mixture configurations of Section~\ref{sec:mixture-qwen25} at the shared 10B-token budget, reported as mean $\pm$ standard deviation over three independent training seeds. Short is the seven-task short-context average; HELMET is the macro-averaged validation score; RULER is the macro-averaged held-out score. The highest value in each column is in bold.}
\label{tab:seed-stability}
\centering
\small
\setlength{\tabcolsep}{8pt}
\renewcommand{\arraystretch}{1.2}
\begin{tabular}{@{}l ccc@{}}
\toprule
Configuration & Short & HELMET & RULER \\
\midrule
dolma3-short & \textbf{60.04\,$\pm$\,0.11} & 49.44\,$\pm$\,0.12 & 77.16\,$\pm$\,0.15 \\
$+$ random-long & 59.81\,$\pm$\,0.04 & 50.72\,$\pm$\,0.10 & 78.61\,$\pm$\,0.15 \\
$+$ \ours{} & 59.79\,$\pm$\,0.06 & \textbf{55.31\,$\pm$\,0.27} & \textbf{81.94\,$\pm$\,0.16} \\
\bottomrule
\end{tabular}
\end{table}

Figure~\ref{fig:short-performance} compares the uncontinued base model with all 6B-token continued-training pipelines from Section~\ref{sec:signals}. Every continued-training configuration has a lower seven-task short-context average than the base model; the \ours{} configuration performs similarly to the other long-context extensions. Table~\ref{tab:short-pertask} reports the per-task short-context scores for the three fixed-mixture configurations of Section~\ref{sec:mixture-qwen25} (Figure~\ref{fig:olmo-production}): the dolma3-short reference, and the two 66/34 mixtures that add random versus \ours{}-selected natural long documents from the same pool. We also include the uncontinued base model (\textsc{Base}, no long-context extension) as a reference. The seven tasks are ARC-Challenge, ARC-Easy, HellaSwag, LAMBADA (OpenAI), LogiQA, PIQA, and WinoGrande, each evaluated with the \texttt{lm-evaluation-harness} \citep{gao2024lmeval}; the final column is their unweighted average. All four configurations stay within 0.3 point on the average: adding selected long documents matches the random-long mixture to within 0.1 point and the dolma3-short reference to within 0.3 point, and remains comparable to the uncontinued base model.

\section{Seed Stability of the Fixed-Mixture Comparison}
\label{app:seed-stability}

We assess seed sensitivity of the fixed-mixture comparison in Section~\ref{sec:mixture-qwen25} by retraining each of the three configurations under three independent seeds and reporting the seven-task short-context average, macro-averaged validation HELMET, and held-out RULER. Table~\ref{tab:seed-stability} gives the mean $\pm$ standard deviation over the three runs. The between-seed standard deviations are small on all three metrics ($\leq 0.27$ HELMET, $\leq 0.16$ RULER, $\leq 0.11$ short-context), so the gaps between configurations are much larger than the seed-level variation: on HELMET and RULER, \ours{} exceeds random-long by $4.59$ and $3.33$ points, respectively, both an order of magnitude above the corresponding standard deviations. Seven-task short-context averages remain within $0.25$ point across all configurations.

\FloatBarrier
\section{Three-Seed Results for the Controlled Qwen2.5-3B Comparison}
\label{app:main-seeds}


Table~\ref{tab:main-seeds} gives the values plotted in Figure~\ref{fig:signals}. For each training run, HELMET is the unweighted average of its five task categories, excluding RULER. We then compute the mean and standard deviation across three run-level scores; RULER is summarized separately. The advantage of \ours{} is preserved in the three-run means. Close rankings among lower-scoring methods are less stable: for example, gzip's HELMET mean is below both \textsc{repack} and the LLM-judge selector, whereas the original reference run put gzip slightly above them.

\begin{table}[!htbp]
\caption{Mean $\pm$ standard deviation over three training seeds for the nine pipelines on Qwen2.5-3B, 128K, 6B tokens.}
\label{tab:main-seeds}
\centering
\small
\begin{tabular}{@{}lcc@{}}
\toprule
Pipeline & HELMET & RULER \\
\midrule
\ours{} & $\mathbf{55.59\pm0.55}$ & $\mathbf{81.31\pm0.14}$ \\
LongFilter & $53.37\pm0.30$ & $79.68\pm0.12$ \\
Prospector & $51.98\pm0.28$ & $78.41\pm0.05$ \\
random-long & $51.05\pm0.56$ & $78.03\pm0.08$ \\
\textsc{repack} & $50.31\pm0.45$ & $76.81\pm0.28$ \\
gzip (mid60) & $50.08\pm0.79$ & $77.68\pm0.26$ \\
LLM-judge (top5) & $50.18\pm0.41$ & $77.46\pm0.02$ \\
PPL (mid40) & $49.62\pm0.58$ & $77.25\pm0.12$ \\
PPL (top40) & $47.41\pm0.41$ & $75.43\pm0.12$ \\
\bottomrule
\end{tabular}
\end{table}


\section{Sensitivity to the Short/Long Mixture Ratio}
\label{app:mixture-ratios}

We compare random-long and \ours{}-selected long data at three short/long mixture ratios in the Qwen2.5-3B, 128K, 10B-token recipe. The 66/34 row reports the rounded summary scores from Section~\ref{sec:mixture-qwen25}, while the other ratios assess sensitivity to the mixture ratio. All ratios below are written in short:long order. The table reports aggregate scores; three-seed uncertainty estimates are available only for the 66/34 setting.

\begin{table}[!htbp]
\caption{Mixture-ratio sensitivity at 128K and 10B tokens on Qwen2.5-3B. Each pair compares random-long and \ours{}-selected long documents. Short is the seven-task short-context average.}
\label{tab:mixture-ratios}
\centering
\small
\setlength{\tabcolsep}{5pt}
\begin{tabular}{@{}lrrrrrr@{}}
\toprule
& \multicolumn{2}{c}{HELMET} & \multicolumn{2}{c}{RULER} & \multicolumn{2}{c}{Short} \\
\cmidrule(lr){2-3}\cmidrule(lr){4-5}\cmidrule(lr){6-7}
Short:long & Random & Selected & Random & Selected & Random & Selected \\
\midrule
66/34 & 50.70 & \textbf{55.30} & 78.60 & \textbf{81.90} & 59.80 & 59.80 \\
50/50 & 51.58 & \textbf{55.42} & 79.02 & \textbf{81.70} & 59.74 & 59.86 \\
34/66 & 49.98 & \textbf{54.35} & 78.18 & \textbf{81.36} & 59.30 & 59.31 \\
\bottomrule
\end{tabular}
\end{table}

The \ours{} mixtures exceed their corresponding random controls on both long-context aggregates at every tested proportion. HELMET gains are 4.60, 3.84, and 4.37 points, and RULER gains are 3.30, 2.68, and 3.18 points, respectively. The seven-task short-context averages are close within each pair. A broader ratio search would be needed to locate an optimal composition.

\Needspace{7\baselineskip}
\section{Qwen3-8B SFT Evaluation Details}
\label{app:qwen3-sft-details}

Tables~\ref{tab:qwen3-sft-longbench-detail} and \ref{tab:qwen3-sft-mrcr-detail} report means and standard deviations over three evaluation seeds for the three Qwen3-8B configurations in Table~\ref{tab:qwen3-main}, each using a fixed SFT model. Continued training uses 256K and 20B tokens, with short:long $=66/34$ in the mixed configurations. SFT sequences are packed to 64K; \ours{} uses a front-half scoring prefix (128K) at $L=256$K.

\paragraph{SFT recipe.}
We fine-tune all configurations on LongMagpie \citep{gao2025longmagpie} for 250 optimization steps with training seed 42. Training uses bf16 and AdamW ($\beta_1{=}0.9$, $\beta_2{=}0.95$, weight decay $0.1$), a learning rate of $1\times10^{-6}$ with cosine decay to $10\%$ of its peak value, a $5\%$ warmup ratio, and gradient clipping at $1.0$. We set \texttt{BSZ}=128 in the training launcher, with a per-device batch size of one, gradient accumulation, and sequence parallelism of size two. Instruction tokens are masked out when computing the loss.

All configurations use evaluation seeds $\{41,42,43\}$ on both benchmarks. All three evaluation runs per configuration and benchmark are included. Comparisons use the configuration means without seed pairing, and the standard deviations quantify evaluation variability of each fixed SFT model. LongBench v2 has $N=503$ and MRCR has $N=1779$ per evaluation run. MRCR overall is summarized directly from the reported metric rather than reconstructed from subset averages.

\begin{table}[!htbp]
\caption{Full LongBench v2 results (\%) for the Qwen3-8B SFT configurations: mean $\pm$ standard deviation over three evaluation seeds. Best means in each row are in bold.}
\label{tab:qwen3-sft-longbench-detail}
\centering
\small
\setlength{\tabcolsep}{5pt}
\renewcommand{\arraystretch}{1.15}
\begin{tabular}{@{}lccc@{}}
\toprule
Metric & dolma3-short & $+$ random-long & $+$ \ours{} \\
\midrule
Overall & 33.20\,$\pm$\,0.40 & 33.87\,$\pm$\,0.31 & \textbf{35.00\,$\pm$\,0.20} \\
Easy & 34.03\,$\pm$\,0.64 & 34.73\,$\pm$\,0.76 & \textbf{35.93\,$\pm$\,1.38} \\
Hard & 32.70\,$\pm$\,0.35 & 33.33\,$\pm$\,0.50 & \textbf{34.40\,$\pm$\,0.60} \\
Short & 42.93\,$\pm$\,0.64 & \textbf{43.33\,$\pm$\,0.55} & 42.60\,$\pm$\,0.82 \\
Medium & 28.67\,$\pm$\,1.21 & 28.67\,$\pm$\,1.10 & \textbf{30.53\,$\pm$\,0.58} \\
Long & 25.93\,$\pm$\,1.62 & 28.43\,$\pm$\,1.93 & \textbf{31.20\,$\pm$\,3.26} \\
\bottomrule
\end{tabular}
\end{table}

\begin{table}[!htbp]
\caption{MRCR overall and needle-subset results (\%) for the Qwen3-8B SFT configurations: mean $\pm$ standard deviation over three evaluation seeds. Best means in each row are in bold.}
\label{tab:qwen3-sft-mrcr-detail}
\centering
\small
\setlength{\tabcolsep}{5pt}
\renewcommand{\arraystretch}{1.15}
\begin{tabular}{@{}lccc@{}}
\toprule
Metric & dolma3-short & $+$ random-long & $+$ \ours{} \\
\midrule
Overall & 24.59\,$\pm$\,0.30 & 24.99\,$\pm$\,0.33 & \textbf{25.61\,$\pm$\,0.29} \\
2-needle & 37.93\,$\pm$\,0.43 & 38.11\,$\pm$\,0.77 & \textbf{39.15\,$\pm$\,0.51} \\
4-needle & 19.99\,$\pm$\,0.60 & 20.61\,$\pm$\,0.24 & \textbf{20.78\,$\pm$\,0.42} \\
8-needle & 15.87\,$\pm$\,0.21 & 16.26\,$\pm$\,0.39 & \textbf{16.90\,$\pm$\,0.21} \\
\bottomrule
\end{tabular}
\end{table}

\paragraph{Results after SFT.}
After SFT, \ours{} scores 35.00 overall on LongBench v2, compared with 33.87 for random-long and 33.20 for dolma3-short. Its Medium and Long scores improve on random-long by 1.87 and 2.77 points, respectively, while its Short score is lower by 0.73 point. For \ours{} on LongBench v2, evaluation-seed variation is greater on Long (standard deviation 3.26) than overall. On MRCR, \ours{} scores 25.61 versus random-long's 24.99 ($+0.62$), with gains on all three needle subsets.

\Needspace{7\baselineskip}
\section{OLMo 3-7B Validation after Supervised Fine-Tuning}
\label{app:olmo-sft}

We provide additional SFT validation for the three OLMo~3-7B continued-training configurations in Table~\ref{tab:model-transfer}: dolma3-short, dolma3-short + random-long, and dolma3-short + \ours{}. The continued-training setting is 256K and 20B tokens, with short:long $=66/34$ in the mixed configurations. The default \ours{} configuration uses a front-half scoring prefix (128K). All three configurations use the LongMagpie SFT recipe described for Qwen3-8B in Appendix~\ref{app:qwen3-sft-details}. Tables~\ref{tab:olmo-sft-longbench} and \ref{tab:olmo-sft-mrcr} give detailed LongBench v2 metrics and MRCR overall and needle-subset results after SFT.

For each fixed SFT model, we report the mean and standard deviation over evaluation seeds $\{41,42,43\}$. LongBench v2 uses $N=503$. MRCR evaluates $N=1780$ examples per run; 620 examples longer than 256K are excluded from scoring. MRCR overall is summarized directly from the reported metric, rather than reconstructed from subset averages. Reported gains are computed from unrounded means.

\begin{table}[!htbp]
\caption{LongBench v2 accuracy (\%) after SFT of OLMo~3-7B: mean $\pm$ standard deviation over three evaluation seeds ($N=503$). Best means in each row are in bold.}
\label{tab:olmo-sft-longbench}
\centering
\small
\setlength{\tabcolsep}{5pt}
\renewcommand{\arraystretch}{1.15}
\begin{tabular}{@{}lccc@{}}
\toprule
Metric & dolma3-short & $+$ random-long & $+$ \ours{} \\
\midrule
Overall & 27.27\,$\pm$\,0.42 & 27.67\,$\pm$\,0.42 & \textbf{29.33\,$\pm$\,0.31} \\
Easy & 27.93\,$\pm$\,2.12 & 29.13\,$\pm$\,1.38 & \textbf{31.40\,$\pm$\,1.21} \\
Hard & 26.90\,$\pm$\,1.23 & 26.80\,$\pm$\,0.85 & \textbf{28.10\,$\pm$\,0.46} \\
Short & 32.03\,$\pm$\,1.96 & 33.87\,$\pm$\,0.55 & \textbf{34.07\,$\pm$\,0.29} \\
Medium & 25.73\,$\pm$\,1.60 & 25.27\,$\pm$\,0.29 & \textbf{27.60\,$\pm$\,0.52} \\
Long & 22.53\,$\pm$\,1.91 & 22.20\,$\pm$\,0.90 & \textbf{25.00\,$\pm$\,0.90} \\
\bottomrule
\end{tabular}
\end{table}

\begin{table}[!htbp]
\caption{MRCR overall and needle-subset accuracy (\%) up to 256K after SFT of OLMo~3-7B: mean $\pm$ standard deviation over three evaluation seeds ($N=1780$). Best means in each row are in bold.}
\label{tab:olmo-sft-mrcr}
\centering
\small
\setlength{\tabcolsep}{5pt}
\renewcommand{\arraystretch}{1.15}
\begin{tabular}{@{}lccc@{}}
\toprule
Metric & dolma3-short & $+$ random-long & $+$ \ours{} \\
\midrule
Overall & 13.02\,$\pm$\,0.18 & 13.73\,$\pm$\,0.27 & \textbf{15.16\,$\pm$\,0.26} \\
2-needle & 17.99\,$\pm$\,0.42 & 18.72\,$\pm$\,1.13 & \textbf{22.03\,$\pm$\,0.83} \\
4-needle & 11.55\,$\pm$\,0.34 & 11.93\,$\pm$\,0.29 & \textbf{12.68\,$\pm$\,0.34} \\
8-needle & 9.54\,$\pm$\,0.31 & 10.54\,$\pm$\,0.47 & \textbf{10.79\,$\pm$\,0.58} \\
\bottomrule
\end{tabular}
\end{table}

\paragraph{Results.}
\ours{} exceeds random-long on both overall scores after SFT. On LongBench v2, it reaches 29.33, compared with 27.67 for random-long, a gain of 1.67 points. On MRCR, it reaches 15.16, compared with 13.73 for random-long, a gain of 1.44 points.

\end{document}

%% file: math_commands.tex
\usepackage{amsmath,amsfonts,bm}
\usepackage{multicol}

\def\eqref#1{equation~\ref{#1}}

\def\1{\bm{1}}

\DeclareMathAlphabet{\mathsfit}{\encodingdefault}{\sfdefault}{m}{sl}
\SetMathAlphabet{\mathsfit}{bold}{\encodingdefault}{\sfdefault}{bx}{n}

